\documentclass[12pt,letterpaper]{article}
\usepackage[a4paper, total={7in, 10in}]{geometry}

\usepackage{helvet}
\usepackage{authblk}
\usepackage[colorlinks=true, linkcolor=black, citecolor=black, urlcolor=blue, filecolor=blue, breaklinks=true, hypertexnames=false]{hyperref}
\usepackage{amsmath}
\usepackage{amssymb}
\usepackage{mathtools}
\usepackage{siunitx}

\usepackage{graphicx}
\usepackage{booktabs}
\usepackage{multirow}
\usepackage{makecell}
\usepackage{tabularx}
\usepackage{array}
\usepackage{adjustbox}
\usepackage{float}
\usepackage{pdflscape}
\usepackage{pdfpages}

\usepackage{xcolor}
\usepackage{colortbl}
\usepackage{multicol}
\usepackage{tcolorbox}
\usepackage{soul}
\sethlcolor{pink!20}

\usepackage[super,comma,sort&compress]{natbib}
\usepackage[T1]{fontenc}
\usepackage[utf8]{inputenc}
\usepackage{newunicodechar}
\newunicodechar{—}{\textemdash}
\newunicodechar{–}{\textendash}
\newunicodechar{‘}{\textquoteleft}
\newunicodechar{’}{\textquoteright}
\newunicodechar{“}{\textquotedblleft}
\newunicodechar{”}{\textquotedblright}
\usepackage{xspace}
\usepackage{longtable}
\usepackage{inconsolata}
\usepackage{enumitem}

\usepackage{xurl}
\usepackage{verbatim}

\newcommand{\Ours}{MedicalAgentsBench\xspace}
\newcommand{\hlc}[2]{\begingroup\sethlcolor{#1}\hl{#2}\endgroup}
\newcommand{\daniel}[1]{#1}
\newcommand{\danielnew}[1]{#1}

\definecolor{ResearchColor}{HTML}{D0E1F9}
\definecolor{MinecraftColor}{HTML}{D5F5E3}
\definecolor{DatabaseColor}{HTML}{FFF9C4}
\definecolor{CodingColor}{HTML}{FFE0B2}
\definecolor{BargainingColor}{HTML}{E1BEE7}
\definecolor{WereWolfColor}{HTML}{F8BBD0}

\makeatletter
\renewcommand{\maketitle}{\bgroup\setlength{\parindent}{0pt}
\begin{flushleft}
  \textbf{\@title}

  \@author
\end{flushleft}\egroup}
\makeatother

\title{MedicalAgentsBench for Complex Medical Reasoning: Comparing Internalized Reasoning Models versus Externalized Agent-based Frameworks}
\date{}

\author[1,8,10]{Yanjun Shao}
\author[2,10]{Xiangru Tang}
\author[3,10]{Jiwoong Sohn}
\author[1,8,10]{Jiapeng Chen}
\author[8]{Yuxuan Liao}
\author[2]{Jiayi Zhang}
\author[2]{Jinyu Xiang}
\author[4]{Fang Wu}
\author[2]{Yilun Zhao}
\author[2]{Chenglin Wu}
\author[5]{Wenqi Shi}
\author[2]{Arman Cohan}
\author[1,2,6,7,8,9,*]{Mark Gerstein~\href{http://orcid.org/0000-0002-9746-3719}{\small ORCID: 0000-0002-9746-3719}}

\affil[1]{Program in Computational Biology \& Biomedical Informatics, Yale University, New Haven, CT 06520, USA}
\affil[2]{Department of Computer Science, Yale University, New Haven, CT 06520, USA}
\affil[3]{Department of Biosystems Science and Engineering, ETH Zurich, 4058 Basel, Switzerland}
\affil[4]{Department of Computer Science, Stanford University, Stanford, CA 94305, USA}
\affil[5]{Department of Bioinformatics, UT Southwestern Medical Center, Dallas, TX 75390, USA}
\affil[6]{Department of Molecular Biophysics and Biochemistry, Yale University, New Haven, CT 06520, USA}
\affil[7]{Department of Statistics \& Data Science, Yale University, New Haven, CT 06520, USA}
\affil[8]{Department of Biomedical Informatics \& Data Science, Yale University, New Haven, CT 06520, USA}
\affil[9]{Lead contact}
\affil[10]{These authors contributed equally}
\affil[*]{Correspondence: pi@gersteinlab.org}

\begin{document}

\maketitle

\section*{SUMMARY}

Complex medical reasoning requires integrating heterogeneous clinical evidence across multiple inference steps. Large language models (LLMs) now approach this through two routes: internalized reasoning and externalized agent scaffolding (frameworks that decompose problems collaboratively amongst multiple LLMs). To determine whether these routes are exclusive or complementary, we introduce MedicalAgentsBench, a filtered benchmark of 862 complex clinical questions drawn from the union of eight medical datasets via difficulty-aware curation and contamination screening. Evaluating three internalized reasoning models (DeepSeek-R1, o1-mini, and o3-mini), seven base models, and nine externalized agent-based methods, we find that internalized and externalized approaches each independently improve performance, and that their benefits compound: the highest accuracy is achieved by layering agent workflows onto an internalized reasoning model (i.e., o3-mini + MDAgents with 35.1\%). Pareto analysis shows this combination dominates the cost--performance frontier; moreover, lightweight optimization on inexpensive models offers an entry point for resource-constrained settings. Our benchmark is at \url{https://github.com/gersteinlab/MedicalAgentsBench}.

\section*{KEYWORDS}

medical reasoning, large language models, multi-agent systems, clinical decision-making, evaluation


\section*{INTRODUCTION}

\begin{figure}[htbp]
\centering
\includegraphics[width=\linewidth]{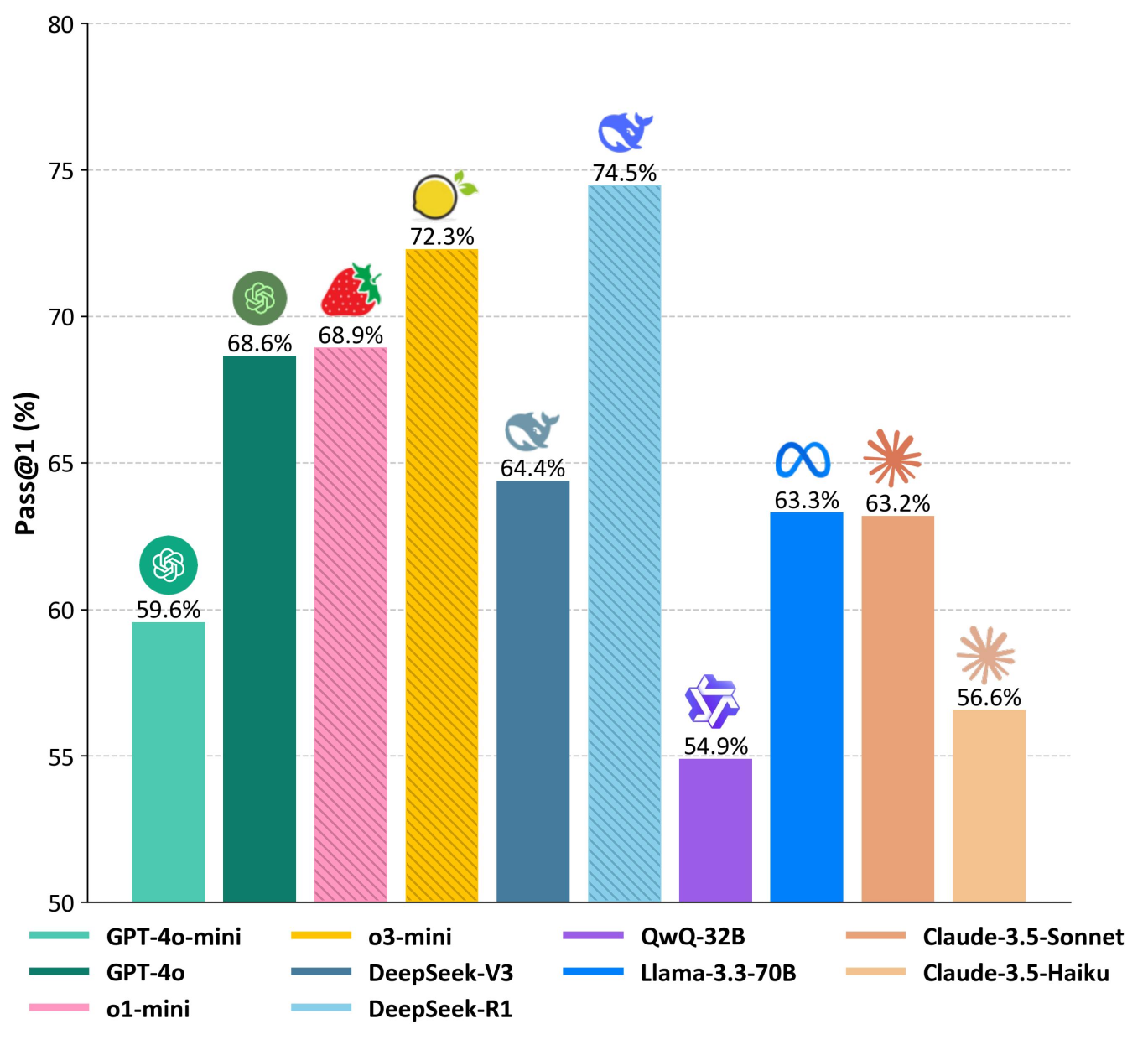}
\caption{\textbf{Performance analysis of large language models on medical tasks.}
Overall \textbf{Pass@1} accuracy comparison across models in zero-shot setting. Hatched bars denote internalized reasoning models.
The score is averaged over ten evaluation subsets (MedQA, PubMedQA, MedMCQA, MedBullets, MMLU, MMLU-Pro, AfrimedQA, MedExQA, and MedXpertQA-R and MedXpertQA-U).}
\label{fig:medagent_teaser_1b}
\end{figure}

\daniel{Clinical reasoning often requires synthesizing heterogeneous evidence (patient history, laboratory values, imaging findings) across multiple inferential steps to arrive at a diagnosis or management decision.} Large language models (LLMs) have demonstrated remarkable capabilities in medical natural language processing, ranging from clinical question answering to diagnostic assistance \cite{singhal2025toward, jin2022biomedical, chen2023systematic, zhou2023survey, gao2024empowering}\daniel{; yet, as illustrated in Figure \ref{fig:medagent_teaser_1b}, even state-of-the-art models (OpenAI o3, GPT-4o, Claude 3.5 Sonnet) continue to struggle with the multi-step reasoning these scenarios demand} \cite{fan2025ai, shi2024medadapter}.

The evolution of medical reasoning evaluation has progressed with numerous specialized datasets. Popular benchmarks like MedQA \cite{jin2021medqa}, PubMedQA \cite{jin2019pubmedqa}, and MedMCQA \cite{pal2022medmcqa} established the foundation with standardized multiple-choice questions from medical licensing exams and PubMed abstracts. The field then expanded to domain-specific resources like clinical notes \cite{pampari2018emrqa} and question summarization \cite{abacha2021overview}, integrating visual elements through datasets like pathology images \cite{he2020pathvqa} and radiology questions \cite{soni2022radqa, bae2023ehrxqa}. To address linguistic and regional diversity, multilingual datasets emerged \citep{vilares2019head, hertzberg2024medqa, olatunji2024afrimed}. The ecosystem further matured with comprehensive collections such as MultiMedQA \cite{singhal2023large}, which combines seven distinct medical datasets.

Research in applying LLMs to medical tasks has progressed through several distinct phases. Initial efforts focused on evaluating general-purpose LLMs in medical contexts \cite{thirunavukarasu2023large, lievin2024can, gilson2023does}. General foundation models \cite{singhal2023large, achiam2023gpt} achieved physician-level performance through prompt engineering, followed by domain-specialized open-source models \cite{han2023medalpaca, wu2024pmc, chen2023meditron}. The field then advanced toward specialized reasoning methods \cite{liu2024medcot, shi2024medadapter} and retrieval-augmented generation \cite{jin2023medcpt, jeong2024improving, xiong2024benchmarking}. More recently, agent-based frameworks such as MedAgents \cite{tang2023medagents} and MDAgents \cite{kim2024mdagents} have shown promise by leveraging collaborative multi-agent systems for complex medical decision-making.


To improve LLM reasoning on such tasks, the community has explored a spectrum of approaches that differ in how reasoning is structured and externalized.
As summarized in Table~\ref{tab:medbench_methods},
existing methods can be broadly categorized into traditional prompting, advanced prompting, discussion-based agent frameworks, and agentic workflow optimization methods, reflecting increasing degrees of reasoning externalization and structural optimization.
Traditional prompting techniques, such as Chain-of-Thought (CoT) and Self-Consistency (SC)~\cite{wei2022chain,wang2022self}, aim to elicit intermediate reasoning steps by prompting models to explicitly verbalize multi-step logical inferences before producing a final answer, with SC further improving robustness through aggregating multiple independently sampled reasoning paths. Advanced prompting methods, including domain-specific approaches such as MedPrompt~\cite{chen2024medprompt}, extend this paradigm by introducing structured self-reflection, diversity, or multi-round prompting within a single model instance, but typically yield only incremental performance gains in complex medical reasoning settings. In contrast, discussion-based agent frameworks such as MedAgents~\cite{tang2023medagents} and MDAgents~\cite{kim2024mdagents} externalize reasoning by coordinating multiple specialized agents that engage in explicit discussion, consultation, and cross-verification, decomposing complex medical reasoning into collaborative subtasks such as hypothesis generation and evidence evaluation. Beyond discussion-based agent frameworks, agentic workflow optimization methods~\citep{zhang2024aflow, xiang2025spo} further automate the design of agent interaction structures by algorithmically optimizing reasoning workflows and control strategies, rather than relying on fixed, hand-designed agent protocols. While agent-based approaches can substantially enhance reasoning performance by externalizing deliberation and verification beyond a single model forward pass, they also introduce increased system complexity and computational overhead.

In parallel, a new class of reasoning models has emerged, including OpenAI o3-mini and DeepSeek R1~\citep{guo2025deepseek}. In this work, we use the term \emph{reasoning models} to refer to LLMs whose parameters are explicitly optimized, via reinforcement learning, chain-of-thought supervision, or distillation, to perform multi-step reasoning internally, rather than relying on external orchestration or tool-based decomposition. \danielnew{A useful overview of this emerging category is provided by~\citet{raschka2025understanding}.}
\danielnew{In effect, these models amortize reasoning into their parameters: training-time optimization shifts more of the cognitive work upstream, so complex inference can be executed within a single model call at deployment time~\cite{guo2025deepseek}.}
Unlike agent frameworks, these models aim to perform complex reasoning within a single forward pass. \daniel{This distinction raises a fundamental and unresolved question: in medical reasoning tasks, do internalized reasoning and externalized agent workflows represent }\danielnew{exclusive, competing paradigms, or are they complementary approaches that can readily be combined?}


Answering this question is hindered by several limitations of existing medical benchmarks. First, as shown in Table \ref{tab:medbench_dataset}, current datasets contain a large proportion of straightforward examination-style questions. On such questions, even base LLMs achieve high accuracy (see Table \ref{tab:model_task_heatmap}, ``Full'' columns), masking meaningful differences between internalized reasoning models and agent frameworks. Second, inconsistent and subjective sampling practices across prior studies \cite{tang2023medagents, kim2024mdagents} prevent reliable cross-method comparison, particularly between approaches that differ substantially in cost and inference structure.
Third, existing benchmarks rarely evaluate performance jointly with computational cost and inference time, despite these factors being central to the design trade-offs of agent-based systems.


To address these challenges, we introduce \Ours{}, a benchmark specifically designed to enable a principled comparison between internalized reasoning models and externalized agent frameworks on complex medical reasoning tasks. 
Unlike existing benchmarks that either focus on general medical knowledge or suffer from ceiling effects, \Ours{} employs a pipeline that: \textbf{(1)} draws from eight diverse established medical datasets (MedQA, PubMedQA, MedMCQA, MedBullets, MMLU, MMLU-Pro, MedExQA, and MedXpertQA); \textbf{(2)} applies difficulty-based filtering to select questions that challenge current models (specifically, questions where fewer than 50\% of tested models provide correct answers); \textbf{(3)} conducts thorough contamination analysis to ensure validity; and \textbf{(4)} incorporates human annotations from medical professionals to verify reasoning depth requirements. Our comprehensive experiments yield several key insights: 
\daniel{\textbf{(a)} both externalized agent scaffolding and internalized reasoning independently improve performance, and their benefits are complementary: layering agent workflows on internalized reasoning models (e.g., o3-mini + MDAgents) yields the highest overall accuracy;
\textbf{(b)} the Pareto frontier of cost--performance trade-offs is dominated by reasoning-model--agent combinations at the high end and by lightweight workflow optimization methods (e.g., SPO) at the low end;
and \textbf{(c)} open-source reasoning models achieve competitive accuracy, though their extended reasoning chains do not necessarily reduce per-sample inference cost.}

Concurrent with our work, several related evaluation efforts have emerged. MedHELM \cite{bedi2026medhelm} provides a holistic evaluation framework for LLMs across a broad range of medical tasks. MedAgentBench \cite{jiang2025medagentbench} introduces a virtual EHR environment for benchmarking clinical agent capabilities. MedAgentBoard \cite{zhu2025medagentboard} benchmarks multi-agent collaboration against conventional methods across diverse medical tasks. MedAgentGym \cite{xu2026medagentgym} proposes a scalable agentic training environment for code-centric reasoning in biomedical data science. Together, these works reflect the growing community interest in rigorous, principled evaluation of LLMs and agent-based systems in medical contexts.

\section*{RESULTS}

\subsection*{Benchmark overview}

\Ours{} is a carefully curated benchmark designed to evaluate complex medical reasoning tasks. Drawing from eight established medical datasets (MedQA~\cite{jin2021medqa}, PubMedQA~\cite{jin2019pubmedqa}, MedMCQA~\cite{pal2022medmcqa}, MedBullets~\cite{chen2024medbullet}, MMLU~\cite{hendrycks2020mmlu}, MMLU-Pro~\cite{wang2024mmlu}, MedExQA~\cite{kim2024medexqa}, and MedXpertQA~\cite{zuo2025medxpertqa}), we systematically construct a challenging subset of 862 questions (average token length: 147.4) that focuses on complex reasoning scenarios. Our multi-stage construction pipeline combines difficulty-aware curation (selecting questions where fewer than 50\% of tested models answer correctly), contamination screening via MELD~\cite{nori2023capabilities}, and human validation of reasoning depth by four medical professionals using a five-level rubric (see Methods for full details). As shown in Table~\ref{tab:model_task_heatmap}, while models achieve high accuracy on the Full set (e.g., GPT-4o: 88.2\% on MedQA), their performance drops significantly on our Hard subset (e.g., GPT-4o: 32.2\%), confirming that the selected questions demand deeper analytical processing and multi-step medical reasoning.



\subsection*{Quantitative evaluations}

\daniel{Most models struggle with our challenging benchmark, with even powerful systems like GPT-4o achieving only 32.2\% accuracy on MedQA and 18.1\% on MedExQA in our Hard subset. This substantial performance drop confirms that our selection criteria effectively identify questions that require advanced reasoning. Amid these challenges, DeepSeek-R1 demonstrates remarkable performance, achieving the highest scores on five datasets (MedMCQA: 30.9\%, MMLU: 43.7\%, MMLU-Pro: 36.8\%, MedExQA: 26.0\%, and MedXpertQA-U: 26.3\%). Similarly, o3-mini excels on two datasets (MedQA: 52.9\% and MedBullets: 50.3\%).}

\daniel{The central finding from our method comparison (Table \ref{tab:method_task_heatmap}) is that externalized and internalized reasoning are complementary. Agent frameworks substantially improve base models: for instance, GPT-4o + AFlow (29.8\% average) outperforms bare GPT-4o zero-shot by a wide margin. Reasoning models independently advance the frontier: bare o3-mini (27.8\%) already surpasses most agent-augmented base-model configurations. Critically, layering agent workflows on top of reasoning models yields further gains: o3-mini + MDAgents achieves the highest overall accuracy (35.1\%), followed by o3-mini + MultiPersona (34.1\%) and o3-mini + CoT (32.7\%), all surpassing bare o3-mini by 5--7 percentage points. This indicates that agents add value even when the underlying model is already strong, likely by introducing diverse reasoning paths and catching errors that a single model pass would miss.}

\daniel{Among individual methods, automated workflow optimization (AFlow) and multi-persona prompting (MultiPersona) achieve the strongest results when applied to base models, while discussion-based agent frameworks (MedAgents, MDAgents) show their greatest value when paired with reasoning models. Traditional prompting gains from CoT-SC over CoT are inconsistent on this hard subset, and domain-specific methods like MedPrompt lack consistency across datasets (see Table \ref{tab:method_task_heatmap} for full per-dataset results).} Additional qualitative error analysis and representative failure cases are provided in Supplemental Note S1.

\subsection*{Benchmark quality assessment}
We assess benchmark quality along two dimensions: reasoning depth to verify cognitive difficulty and data contamination to ensure fair and reliable model comparison.
\subsubsection*{Reasoning depth analysis}
\label{sec:reasoning_depth_analysis}

\begin{figure}[htbp]
\centering
\includegraphics[width=\linewidth]{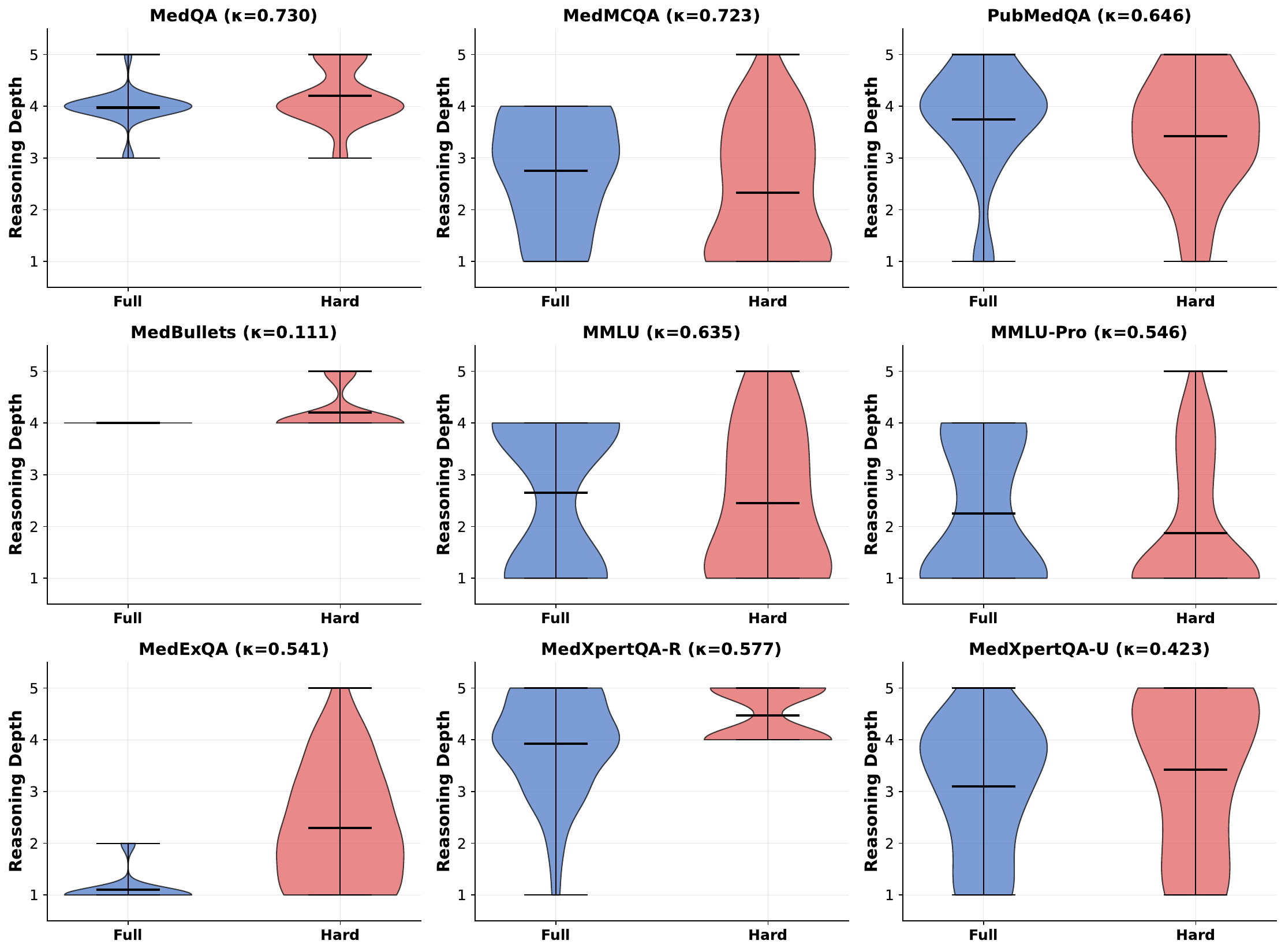}
\caption{\textbf{Reasoning depth distribution by dataset split.}
Violin plots comparing the density of reasoning depth ratings (Levels 1--5) between the complete source datasets (Full) and the curated \Ours{} subset (Hard). $\kappa$ values indicate inter-annotator agreement for each domain. Full details of annotation can be found in the human validation protocol in Methods.}
\label{fig:reasoning_depth}
\end{figure}

To directly assess whether \Ours{} emphasizes multi-step clinical reasoning rather than superficial difficulty or obscure knowledge retrieval, we conducted an expert-annotated analysis of reasoning depth using a standardized five-level rubric detailed in the human validation protocol in Methods. Four medical students in their clinical years independently annotated a stratified sample of 20 questions per dataset (10 from the Full split and 10 from the Hard split), covering all eight source datasets across nine evaluation subsets (180 questions total). This design enables a direct comparison of the cognitive demands imposed by the original datasets versus the curated benchmark. Figure~\ref{fig:reasoning_depth} reveals a clear structural shift in cognitive demands between the two splits. In the full distributions of generalist benchmarks such as MMLU and MMLU-Pro, reasoning depth is bottom-heavy, with a predominance of Level 1 and Level 2 questions solvable via direct fact retrieval or single-step deduction. In contrast, the Hard subset exhibits a pronounced shift toward Levels 4 and 5, corresponding to multi-hop clinical reasoning that requires synthesizing heterogeneous evidence or resolving competing hypotheses.


This pattern is consistent across diverse medical domains and is not driven by a small number of outlier datasets. For example, in MedXpertQA, the Hard subset effectively removes direct-recall questions and concentrates on scenarios that require integrating patient history, vital signs, and diagnostic findings to arrive at a diagnosis or management decision. Even for case-based datasets such as MedBullets, which already operate at relatively high reasoning levels, the Hard split further emphasizes Level 5 reasoning involving conflicting clinical evidence or subtle contraindications.
Inter-annotator agreement is substantial (overall Fleiss' $\kappa = 0.62$, Supplemental Table S3), indicating that these distinctions in reasoning depth are clinically meaningful and reproducible rather than subjective.

\subsubsection*{Data contamination analysis}
\label{sec:data_contamination_analysis}

\begin{figure}[htbp]
\centering
\includegraphics[width=\linewidth]{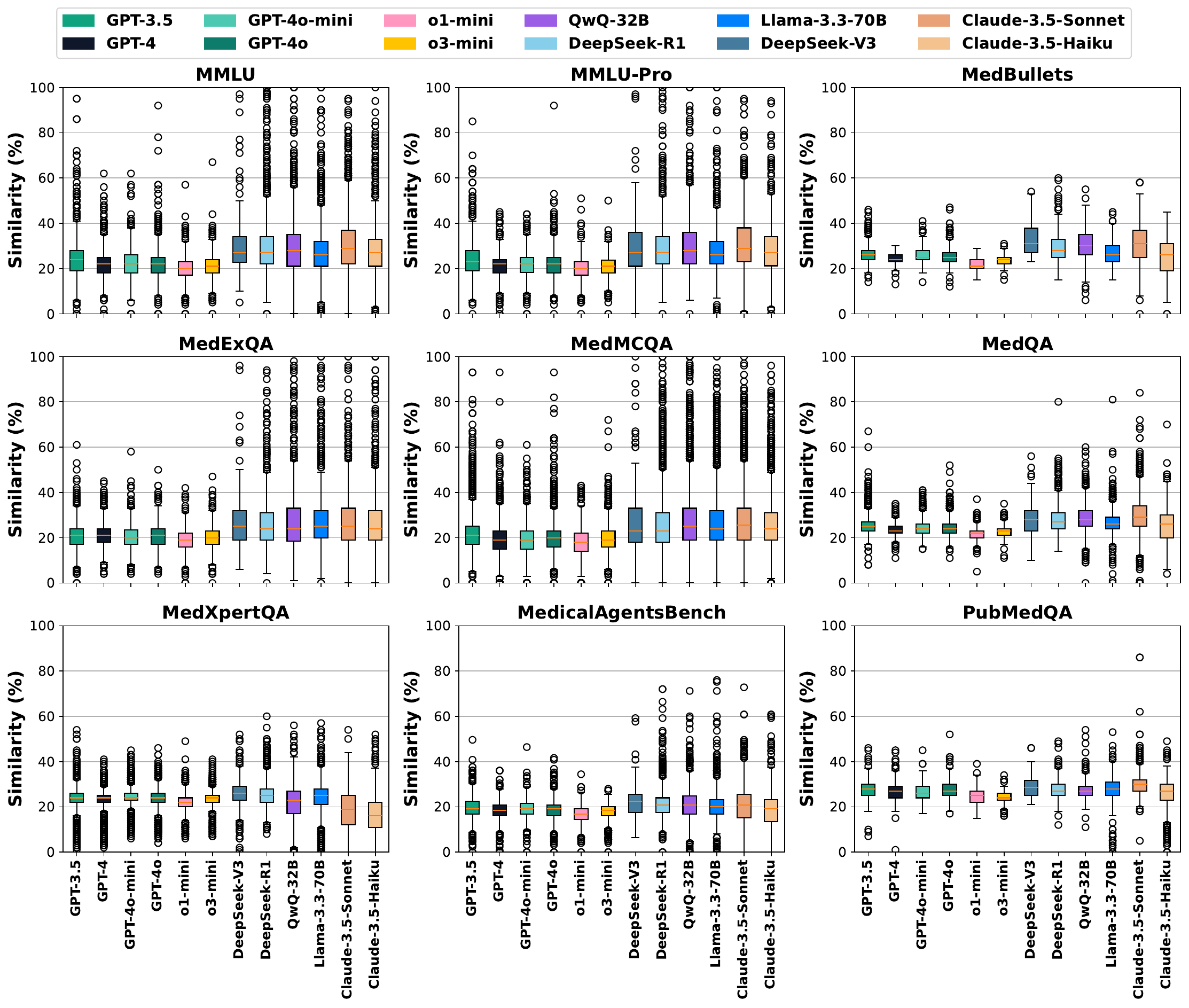}
\caption{\textbf{Data contamination analysis across medical question-answering datasets using MELD.}
The boxplots display similarity percentages between model-generated text and original question text, with higher values potentially indicating memorization of training data.
Lower similarity scores suggest minimal data contamination, while higher values may indicate potential contamination in model training data.}
\label{fig:data_contamination}
\end{figure}

\daniel{We assessed potential data contamination using the MELD (Memorization Effects Levenshtein Detector) framework~\cite{nori2023capabilities}. MELD splits each question into two halves: the first half is provided as context, and the model is prompted to generate the second half. The Levenshtein distance ratio between the generated output and the original unseen portion quantifies memorization; higher similarity indicates that the model has likely encountered the question during training. This approach is particularly well-suited to multiple-choice benchmarks, where verbatim reproduction of question text, rather than answer choices, serves as the primary indicator of training exposure (Figure~\ref{fig:data_contamination}). MELD has high precision but unknown recall: a detected match strongly indicates memorization, but absence of a match does not guarantee clean data.}

\daniel{The analysis reveals a clear divide between commercial and open-source models. OpenAI models show consistently low similarity scores (median 20--25\%), suggesting minimal verbatim memorization. In contrast, several open-source models (DeepSeek-R1, QwQ-32B, Llama-3.3-70B) exhibit substantially higher scores, with some reproducing question texts at over 95\% character-level accuracy on MMLU and MMLU-Pro (60--90\% similarity range). Importantly, our Hard subset demonstrates significantly lower MELD scores across all models than the corresponding Full sets, confirming its utility for more reliable cross-model comparison. A manual review of high-similarity cases in MedMCQA and MedExQA attributed some elevated scores to shared standard medical terminology rather than direct data leakage. }\danielnew{Given these contamination concerns, our primary method comparisons (Table \ref{tab:method_task_heatmap}) focus on OpenAI models (GPT-4o-mini, GPT-4o, o1-mini, o3-mini), which exhibit the lowest contamination risk and therefore provide the most reliable basis for comparing reasoning strategies. Search-based optimization methods (SPO and AFlow) were evaluated only for GPT-4o-mini and GPT-4o.}

\subsection*{Cost-performance analysis}

Our evaluation examines both model architecture efficiency and reasoning method effectiveness through a cost-performance lens. We analyze these dimensions separately using two complementary visualizations.

\subsubsection*{Base model cost-efficiency}

\begin{figure}[htbp]
\centering
\includegraphics[width=\linewidth]{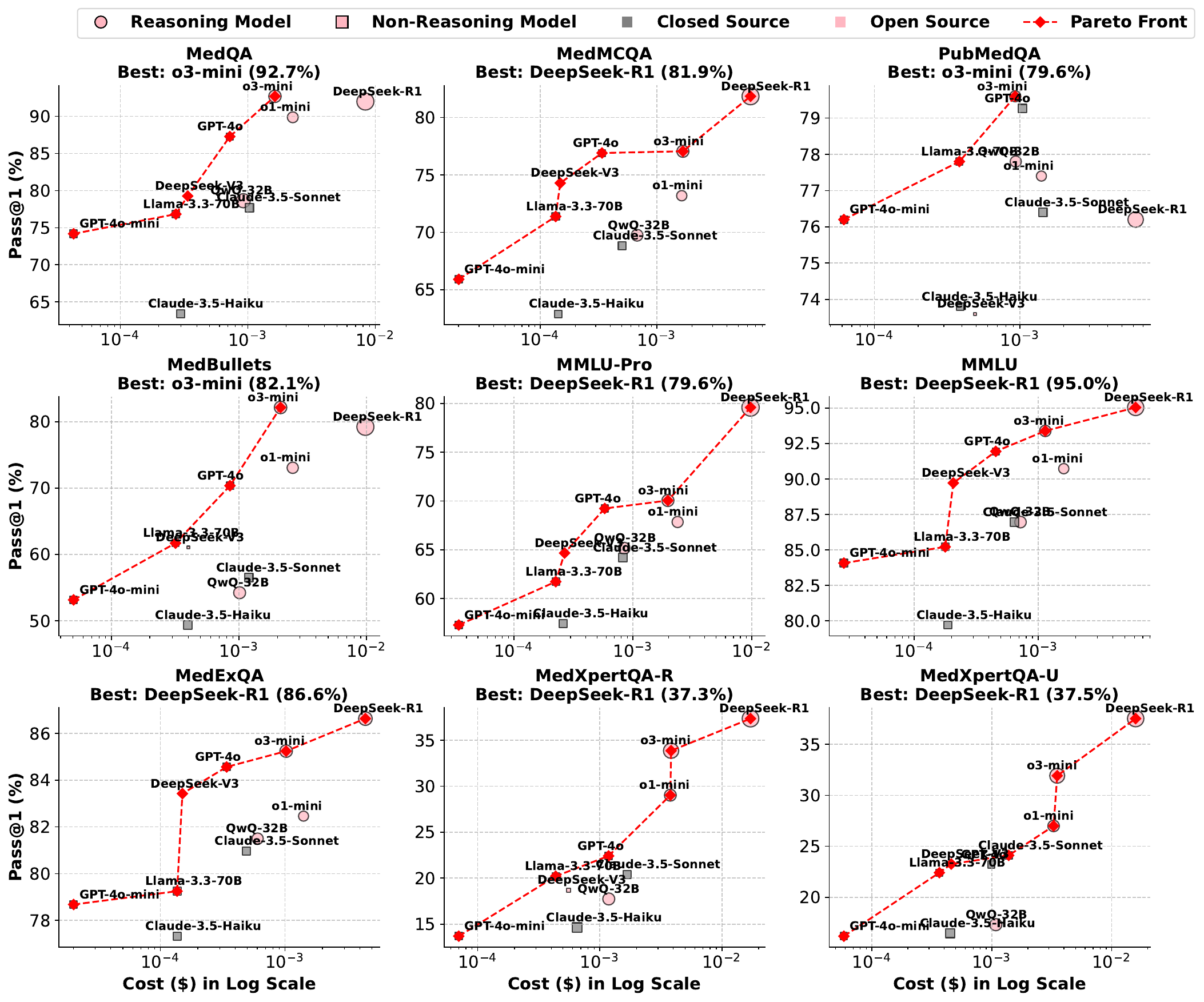}
\caption{\textbf{Cost-performance analysis across medical datasets, comparing open and closed-source language models.}
Each subplot shows the ten base models for one dataset, plotting Pass@1 accuracy (\%) against inference cost per sample (USD, log scale). Circles denote internalized reasoning models and squares denote non-reasoning models; pink markers indicate open-source models and gray markers indicate closed-source models. The red dashed line traces the empirical Pareto frontier, and each subplot title reports the highest-accuracy model for that dataset.}
\label{fig:pareto_front}
\end{figure}

\danielnew{As shown in Figure \ref{fig:pareto_front}, the ten base models occupy markedly different positions in accuracy--cost space across the eight datasets. The marker encoding and Pareto frontier are described in the figure caption; here, we focus on the substantive pattern. In each panel, the red dashed frontier identifies non-dominated models that achieve the best accuracy attainable at a given inference cost, whereas points below the frontier are strictly inefficient relative to another model on the same dataset. DeepSeek-R1 and o3-mini recur on or near this frontier across many datasets, indicating that they frequently define the strongest empirical trade-offs among the base models.}

\danielnew{The frontier nevertheless varies by task. MedQA and MedBullets show steep gains as inference cost increases, whereas PubMedQA exhibits a flatter profile with more modest returns to expensive models. Across several of the more challenging datasets, internalized reasoning models occupy stronger positions than non-reasoning models at comparable cost levels, suggesting that explicit training for multi-step reasoning can improve deployment-time efficiency as well as accuracy.}
\daniel{Open-source models demonstrate competitive performance. On the Full test set, DeepSeek-R1 achieves the highest accuracy among all models on several benchmarks, including MedMCQA (82.2\%) and MMLU-Pro (79.7\%), rivaling or surpassing closed-source alternatives. However, its extended reasoning chains result in higher per-sample inference costs than compact closed-source models such as o3-mini, illustrating a trade-off between reasoning depth and cost efficiency.}

\subsubsection*{Reasoning method cost-efficiency}
\begin{figure}[htbp]
\centering
\includegraphics[width=\linewidth]{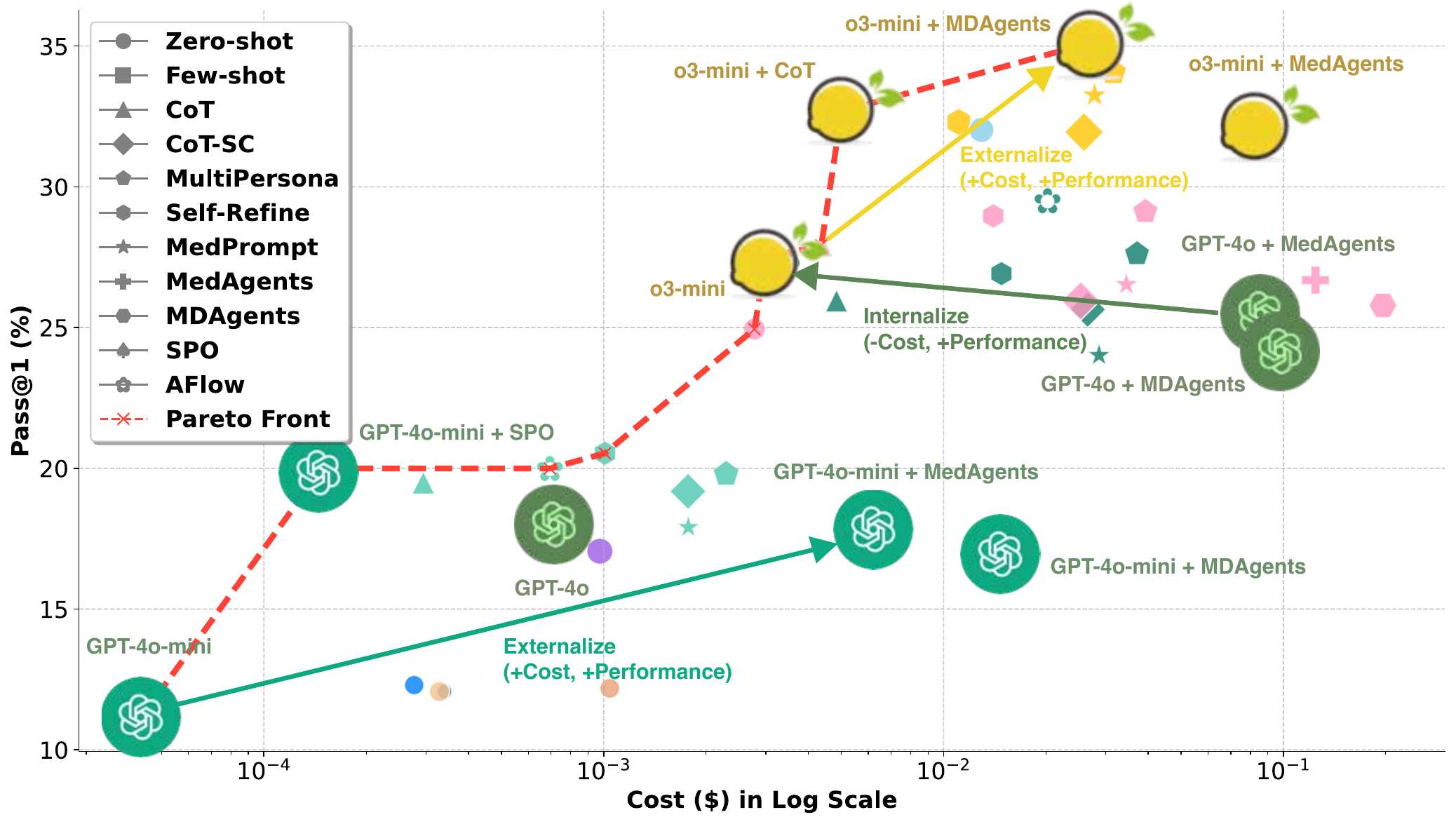}
\caption{\textbf{Cost-performance analysis of reasoning methods on \Ours{}.}
Pass@1 accuracy versus inference cost per sample (log scale). Different markers represent reasoning methods; base-model logo colors follow Figure~1; marker sizes indicate inference time. The red dashed line traces the Pareto frontier of non-dominated configurations. Annotated arrows highlight two distinct mechanisms: \emph{Externalizing} reasoning (adding agent scaffolding to a fixed model) increases both accuracy and cost, while \emph{Internalizing} reasoning (switching to a reasoning model) improves accuracy while reducing inference cost.}
\label{fig:medagent_pareto}
\end{figure}

\daniel{Figure \ref{fig:medagent_pareto} extends our analysis to the full space of model--method combinations. The Pareto frontier, together with the annotated transitions in the figure, reveals two distinct mechanisms for improving medical reasoning, with opposite cost signatures.}

\daniel{\textbf{Externalizing reasoning trades cost for performance.} Adding agent scaffolding to a fixed base model consistently improves accuracy but increases inference cost. For instance, moving from bare GPT-4o-mini (11.0\%) to GPT-4o-mini + SPO (20.1\%) nearly doubles performance at a modest cost increase. However, not all externalization is equally efficient: GPT-4o-mini + MedAgents (18.3\%) costs far more than SPO yet achieves lower accuracy, illustrating that heavier agent scaffolding can yield diminishing or even negative returns on weak base models. The same pattern holds at the top of the frontier: o3-mini zero-shot (27.8\%) improves to o3-mini + CoT (32.7\%) and then to o3-mini + MDAgents (35.1\%), each step adding both accuracy and cost.}

\daniel{\textbf{Internalizing reasoning improves performance while reducing inference cost.} Switching from an agent-augmented base model to a bare reasoning model can simultaneously increase accuracy and decrease per-sample inference cost. The most striking example is the transition from the cluster of GPT-4o + agent configurations (25--30\% accuracy at 10--100$\times$ higher inference cost) to bare o3-mini (27.8\%): the reasoning model achieves comparable or higher accuracy at an order of magnitude lower inference cost. This reflects the efficiency of internalized reasoning at deployment time, where multi-step inference is compiled into the model's parameters rather than orchestrated through repeated API calls. We note, however, that this analysis considers only inference cost; the training cost required to develop reasoning capabilities (e.g., reinforcement learning, chain-of-thought distillation) is not publicly disclosed by model providers and may be substantial.}

\daniel{These two mechanisms are complementary and together define the Pareto frontier. The optimal strategy traces a path that first externalizes reasoning on cheap models (GPT-4o-mini + SPO), then internalizes by upgrading to a reasoning model (o3-mini), and finally externalizes again by layering agents on the reasoning model (o3-mini + MDAgents). Agent frameworks applied to non-reasoning models (e.g., GPT-4o + MedAgents) fall well below this frontier, as their high inference cost is not justified by commensurate accuracy gains. Practitioners should therefore view internalized reasoning as the primary investment for improving accuracy per dollar, with externalized agent scaffolding as an additive enhancement best applied on top of already-strong reasoning models.}

\section*{DISCUSSION}

\daniel{The central finding of this study is that internalized reasoning and externalized agent scaffolding are not substitutes but complements. Agents lift base-model accuracy (GPT-4o to GPT-4o + MedAgents), internalized reasoning models do the same at lower inference cost (GPT-4o + agents to o3-mini), and agents on top of internalized reasoning models push accuracy further still (o3-mini to o3-mini + MDAgents, 35.1\%). Neither approach subsumes the other. Internalized reasoning models internalize multi-step inference through training-time optimization, while agent frameworks introduce diverse viewpoints and iterative error correction at inference time. The two mechanisms operate on different failure modes, which explains why their gains compound rather than overlap.}

\daniel{The Pareto frontier analysis further clarifies when each approach is most valuable. At the budget-conscious end, lightweight workflow optimization methods such as SPO applied to inexpensive base models (e.g., GPT-4o-mini + SPO) provide the most cost-efficient entry point. At the high-performance end, internalized reasoning models combined with agent methods (o3-mini + CoT, o3-mini + MDAgents) dominate the frontier. Discussion-based frameworks (MedAgents, MDAgents) show their greatest value when paired with internalized reasoning models, where the additional inference cost is justified by meaningful accuracy gains.} Our error analysis (Supplemental Note S1) confirms this mechanistically: multi-agent aggregation amplifies shared biases when agents lack diverse reasoning priors, whereas structured self-critique and role-specialized disagreement can recover from localized errors.

\daniel{For deployment, these results suggest a simple decision rule. If the budget allows an internalized reasoning model, use one; agents on top will further improve accuracy. If not, lightweight workflow optimization (e.g., SPO on GPT-4o-mini) offers the best accuracy per dollar. In either regime, the choice is not ``agents or internalized reasoning models'' but ``how much of each.'' }\danielnew{This trade-off is especially salient in healthcare, where privacy requirements and local infrastructure constraints may preclude sending sensitive clinical data to frontier proprietary systems~\cite{thirunavukarasu2023large,peng2023study}. In practice, the key question is which combination of internalized reasoning models and externalized workflows best satisfies privacy, compute, latency, and accuracy requirements.}

More broadly, \Ours{} demonstrates the value of decision-oriented benchmarking that jointly evaluates accuracy, cost, and reasoning quality. By filtering for genuinely difficult questions and screening for data contamination, our benchmark exposes meaningful performance differences that standard medical QA datasets obscure through ceiling effects.

\subsection*{Limitations of the study}

Our benchmark draws exclusively from multiple-choice QA datasets derived from educational resources, which do not capture the full complexity of clinical practice (e.g., longitudinal patient management, ambiguous presentations, multi-modal evidence integration). Validating these findings on real-world clinical decision-making tasks remains an important next step.

The 50\% accuracy threshold used for difficulty filtering introduces a potential circularity: the ``hard'' subset is defined relative to the specific models evaluated. We mitigate this through architectural diversity in the model pool and independent human validation of reasoning depth, but future work should explore task-intrinsic difficulty measures that do not depend on model performance.

\danielnew{Our cost comparisons focus on inference-time expense. For proprietary internalized reasoning models such as o3-mini, the training cost required to produce these reasoning capabilities is not publicly disclosed, making it difficult to compare the full end-to-end economic trade-off between internalizing reasoning during training and externalizing it through agent scaffolding at inference time.}

Finally, our evaluation covers a fixed set of nine agentic methods. More sophisticated strategies, such as step-wise verification, heterogeneous model ensembles, and dynamic agent collaboration, remain unexplored and could alter the comparative landscape between internalized and externalized reasoning.

\section*{METHODS}

\subsection*{Evaluation tasks and datasets}

\Ours{} is a carefully curated benchmark designed to evaluate complex medical reasoning tasks. We draw from eight established medical datasets, described below.

\paragraph{MedQA}
MedQA \cite{jin2021medqa} is a large-scale open-domain medical question answering dataset collected from professional board exams such as the USMLE. In our benchmark, we use 1{,}273 English multiple-choice questions that test a broad range of clinical and foundational medical knowledge.

\paragraph{PubMedQA}
PubMedQA \cite{jin2019pubmedqa} is a biomedical question answering dataset constructed from PubMed abstracts to evaluate reasoning over primary research literature. We use 500 expert-annotated yes/no/maybe questions, where each item pairs a research question with its corresponding abstract.

\paragraph{MedMCQA}
MedMCQA \cite{pal2022medmcqa} is a large-scale multiple-choice dataset built from Indian medical entrance exams (AIIMS and NEET PG), covering many subjects and topics. Our benchmark includes 2{,}816 questions sampled from MedMCQA, focusing on diverse real-world exam-style problems.

\paragraph{MedBullets}
MedBullets \cite{chen2024medbullet} comprises USMLE Step 2/3-style clinical questions sourced from the MedBullets online learning platform. We use 308 challenging case-based questions, each designed to simulate realistic clinical decision-making scenarios with expert-authored explanations.

\paragraph{MedExQA}
MedExQA \cite{kim2024medexqa} is a benchmark designed to evaluate medical question answering with a focus on explanation quality across underrepresented specialties. We extract 935 questions spanning Audiology, Nutrition, Occupational Therapy, Physical Therapy, and Speech-Language Pathology, each paired with multiple expert explanations.

\paragraph{MedXpertQA}
MedXpertQA \cite{zuo2025medxpertqa} is a recent benchmark targeting expert-level medical knowledge and advanced clinical reasoning across multiple specialties and body systems. We include 2{,}450 text-based questions from MedXpertQA, covering both understanding-oriented and reasoning-intensive scenarios.

\paragraph{MMLU}
The Massive Multitask Language Understanding benchmark \cite{hendrycks2020mmlu} includes several medically relevant subjects. Following the Med-PaLM setting, we retain only the medical fields clinical\_knowledge, professional\_medicine, college\_medicine, medical\_genetics, anatomy, and college\_biology, yielding 1{,}089 questions.

\paragraph{MMLU-Pro}
MMLU-Pro \cite{wang2024mmlu} is an enhanced, more challenging extension of MMLU that emphasizes reasoning and harder distractors. We include only subjects whose source (src) belongs to the health category, specifically clinical\_knowledge, professional\_medicine, college\_medicine, medical\_genetics, nutrition, human\_aging, anatomy, and virology, resulting in 818 questions.

These source datasets vary significantly in size (from 308 to 2,816 questions), average token length (18.7 to 316.1), and number of options (3 to 10), providing diverse evaluation contexts (Table~\ref{tab:medbench_dataset}).

\subsection*{Benchmark construction}

\begin{figure}[htbp]
\centering
\includegraphics[width=\linewidth]{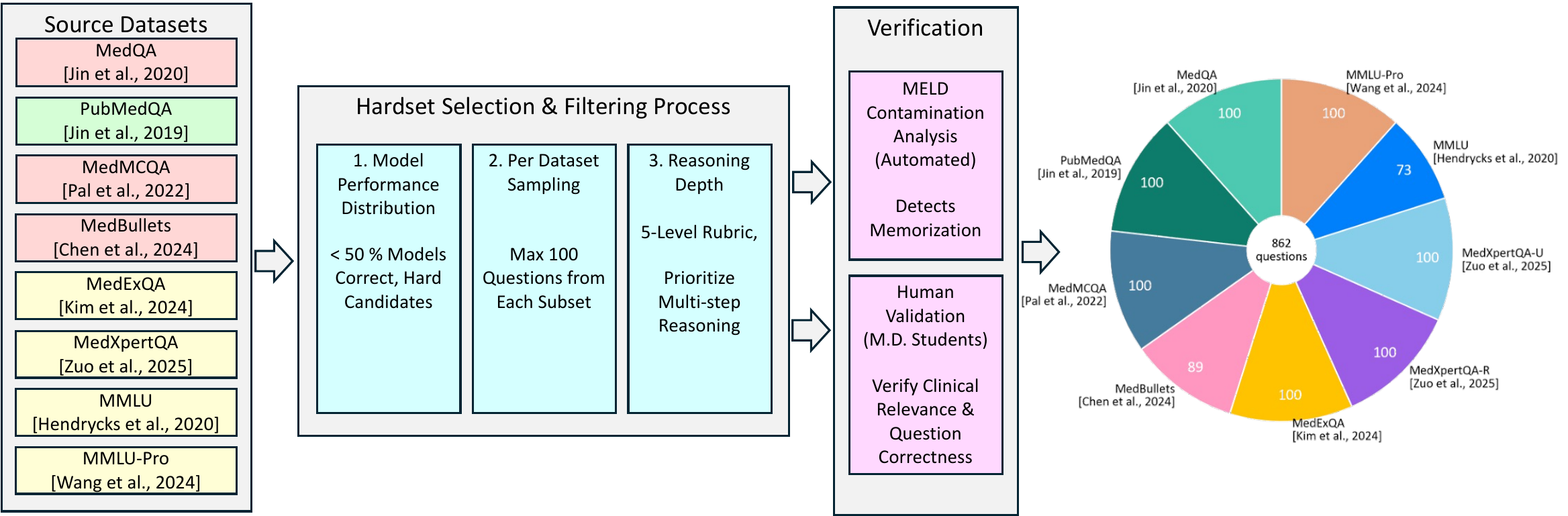}
\caption{\textbf{Hard-set construction pipeline for \Ours{}.}
Questions are aggregated from eight established medical QA datasets and filtered via a three-stage selection process based on model performance, per-dataset sampling, and reasoning depth.
The resulting set is verified using automated contamination analysis and human validation by M.D. students.
The final benchmark contains 862 curated hard questions, with per-dataset composition shown on the right.}
\label{fig:hardset_pipeline}
\end{figure}
\begin{figure}[htbp]
\centering
\includegraphics[width=\linewidth]{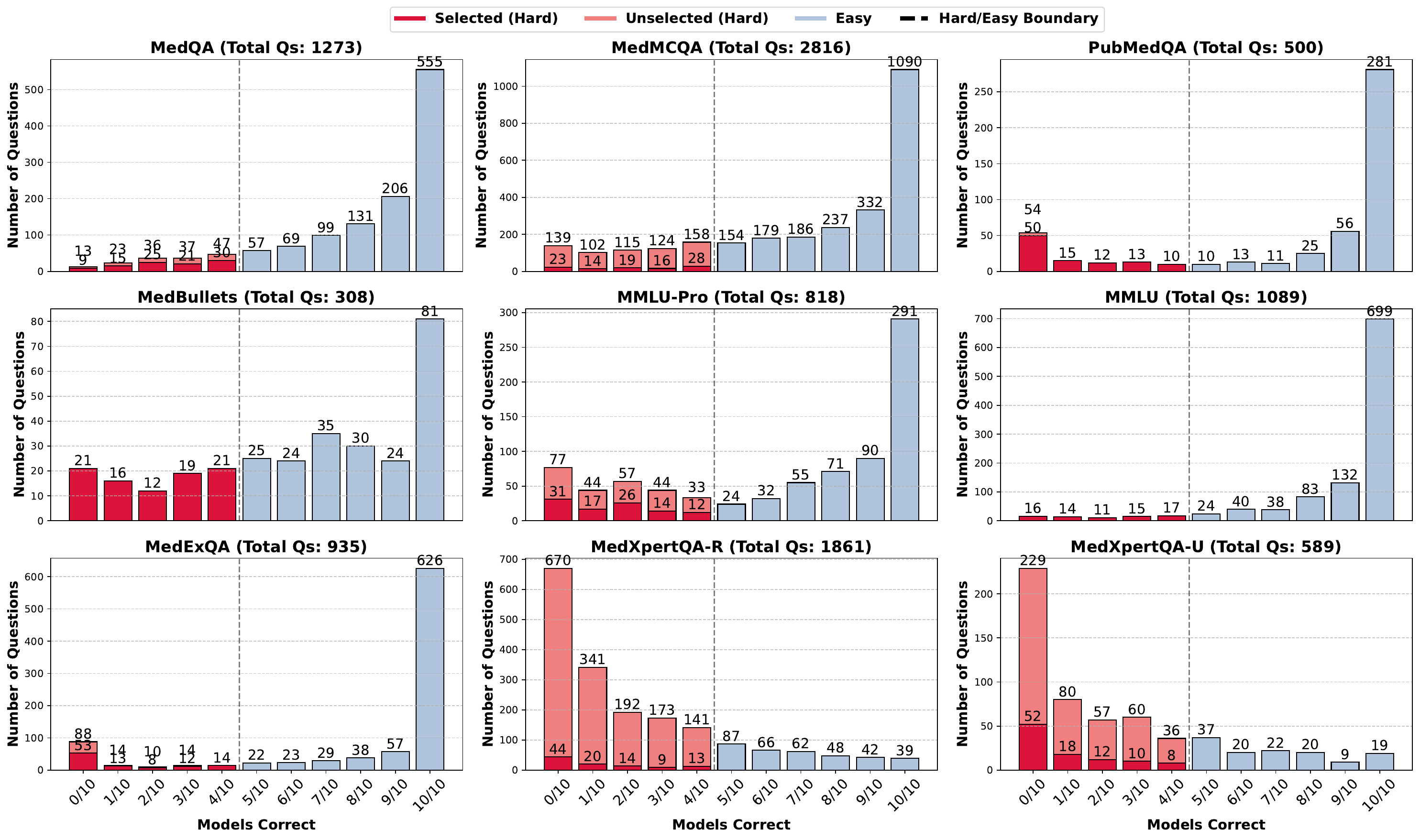}
\caption{\textbf{Distribution of model performance across eight medical datasets.}
MedQA, MedMCQA, PubMedQA, MedBullets, MMLU-Pro, MMLU, MedExQA, and MedXpertQA. Each subplot shows the number of questions answered correctly by different proportions of models (x-axis: k/N, where k is the number of correct models and N is the total number of models). Questions are categorized as either hard (left of the dashed line, $<$ 50\% of models correct) or easy (right of the dashed line, $\geq$ 50\% of models correct), with selected questions highlighted in darker shades. The total question count for each dataset is indicated in the subplot titles.}
\label{fig:answer_correct_distribution}
\end{figure}

As illustrated in Figure~\ref{fig:hardset_pipeline}, we construct \Ours{} by aggregating questions from multiple medical QA datasets and applying a multi-stage filtering and verification process to ensure difficulty, reasoning depth, and clinical validity. Our hard-set selection process is based on three key criteria:

\paragraph{Model performance distribution} As visualized in Figure~\ref{fig:answer_correct_distribution}, we analyze the proportion of models that correctly answer each question ($k/N$ ratio). Questions where less than 50\% of models provide correct answers (left of the dashed line in Figure~\ref{fig:answer_correct_distribution}), are categorized as hard candidates. This ensures our benchmark focuses on truly challenging questions that current models struggle with.

\paragraph{Dataset sampling strategy} We carefully balanced our benchmark to ensure comprehensive coverage across various source datasets. Our Hard subset includes precisely 100 questions each from MedQA, PubMedQA, MedMCQA, MMLU-Pro, and MedExQA. From MedXpertQA, we incorporated both its Reasoning (MedXpertQA-R) and Understanding (MedXpertQA-U) subsets (100 questions each), which were annotated in the original paper to distinguish between questions requiring complex clinical reasoning versus those primarily testing medical knowledge. Additionally, we included 89 questions from MedBullets and 73 questions from MMLU. This sampling distribution maintains proportional representation across the source datasets while ensuring sufficient sample size to evaluate model performance across different medical question types and their associated reasoning requirements.

\paragraph{Reasoning depth} We prioritize questions that require multi-step reasoning processes, where solving the problem involves a sequence of logical deductions or inferences rather than simple fact retrieval. This type of reasoning is typically demonstrated in long chain-of-thought approaches, where models must work through several intermediate steps to reach a conclusion. To systematically assess reasoning depth, we developed a five-level classification rubric and engaged medical professionals to annotate each question (see the human validation protocol below for the complete rubric and validation details). The need for such complex reasoning is evidenced by the significant performance gap between base models and agent-based approaches. As shown in Table~\ref{tab:model_task_heatmap}, while models achieve high accuracy on the Full set (e.g., GPT-4o: 88.2\% on MedQA), their performance drops significantly on our Hard subset (e.g., GPT-4o: 32.2\% on MedQA-Hard), confirming that these selected questions demand deeper analytical processing and medical reasoning capabilities.

In summary, we construct MedicalAgentsBench through a multi-stage pipeline designed to ensure question difficulty, diversity, and evaluation validity. Specifically, we first evaluate each candidate question across multiple model architectures (as shown in Table~\ref{tab:model_task_heatmap}) to ensure architecture-independent difficulty. Second, we apply a difficulty-aware dataset sampling strategy to balance coverage across source benchmarks. Third, we incorporate human annotations to verify that selected questions require multi-step medical reasoning rather than surface-level recall. Finally, we conduct a data contamination analysis using MELD (Memorization Effects Levenshtein Detector)~\cite{nori2023capabilities} to identify questions that may have been memorized during model training. This analysis involves splitting each test item (including the question text but excluding answer options) into two halves, providing models with only the first half, and measuring the similarity between their generated continuations and the original second half. By filtering out potentially memorized items, this step safeguards the fairness and validity of the benchmark for evaluating genuine medical reasoning capabilities.
As shown in Figure~\ref{fig:data_contamination}, the Hard subset selected for our \Ours{} benchmark consistently demonstrates lower similarity scores (20-40\%) compared to the original datasets, suggesting our filtering process effectively selects questions that test genuine reasoning rather than memorization.
Complete details of our MELD methodology and its limitations are discussed in the data contamination analysis under Results. Finally, four medical professionals (M.D. students) review the final question set to verify clinical relevance and reasoning complexity through the validation protocol described below. The resulting \Ours{} benchmark contains 862 questions with an average token length of 147.4.


\subsection*{Human validation protocol}
\label{sec:human_validation}

To ensure the quality and clinical validity of benchmark questions, we conducted a human validation study using four medical students in their clinical years with formal clinical training as annotators. The validation process served two primary objectives: assessing the level of reasoning depth required to answer each question and verifying the correctness and clinical relevance of the questions. Each annotator received standardized training materials and participated in a calibration session, during which they jointly reviewed and discussed 10 representative sample questions to ensure consistent interpretation and application of the evaluation criteria. Following calibration, annotators independently reviewed assigned question subsets, with overlapping samples included to enable estimation of inter-annotator agreement.

\paragraph{Reasoning depth assessment}
For reasoning depth assessment, we developed a systematic rubric that defines reasoning depth as the minimum number of distinct cognitive inference steps required to arrive at the correct answer. Each step corresponds to a discrete inference, calculation, or clinical judgment that builds upon previously derived information. Supplemental Table S1 summarizes the five-level reasoning depth classification system and provides representative clinical examples for each level. Annotators independently reviewed each question, documented the required reasoning steps, and assigned depth scores according to this rubric. To enable reliable estimation of inter-annotator agreement, we sampled 20 questions from each dataset (10 from the original dataset and 10 from the curated hard subset) for comparative annotation by multiple reviewers.

Inter-annotator reliability was quantified using Fleiss' $\kappa$, which measures agreement beyond chance among multiple raters. Specifically, for each item, we compute the proportion of rater pairs that assign the same category and average this quantity across items to obtain the observed agreement $\bar{P}$. The expected agreement $\bar{P}_e$ is computed from the squared category frequencies averaged across all ratings. Fleiss' $\kappa$ is then defined as
\[
\kappa = \frac{\bar{P} - \bar{P}_e}{1 - \bar{P}_e}.
\]
Across all datasets, the resulting inter-annotator agreement is $\kappa = 0.62$ (Supplemental Table S3), which is commonly considered substantial agreement for complex cognitive annotation tasks.


\paragraph{Question validity assessment}
Concurrent with reasoning depth assessment, annotators evaluated question validity using strict exclusion criteria.
Questions were flagged for removal if they contained wrong answers contradicting current medical guidelines, lacked sufficient clinical context for definitive answers, tested outdated or clinically irrelevant knowledge that could be reasonably interpreted multiple ways.
Supplemental Table S2 summarizes our question exclusion criteria and the review process.


Our quality control protocol implemented several measures to ensure annotation consistency.
Following the initial calibration phase, any question flagged by at least one annotator underwent group review.
Questions were removed from the final dataset if two or more annotators agreed on the validity of the concern, with all exclusions documented with a specific rationale.
Of the initial candidate questions, 21 questions were removed due to validity concerns: 5 for wrong answers, 5 for poor clinical context, and 11 for clinical irrelevance.

\subsection*{Large language models}


We conduct a comprehensive evaluation of both base models and agentic reasoning methods across our \Ours{} benchmark. Our experiments follow a standardized protocol to ensure fair comparison, with consistent prompting strategies and evaluation metrics across all models and methods.

\paragraph{Representative LLMs and their versions}
For base model comparison, we evaluate both closed-source models (GPT-4o, GPT-4o-mini, Claude-3.5-Sonnet, Claude-3.5-Haiku, o1-mini, and o3-mini) and open-source alternatives (DeepSeek-V3, DeepSeek-R1, Llama-3.3-70B, and QwQ-32B). Specific model versions were accessed between January and February 2025. For closed-source models via OpenAI and Anthropic APIs, we used gpt-4o-2024-08-06, gpt-4o-mini-2024-07-18, claude-3-5-sonnet-20241022, claude-3-5-haiku-20241022, o1-mini-2024-09-12, and o3-mini-2025-01-31. For open-source models hosted on Together AI (\url{https://www.together.ai/}), we used deepseek-ai/DeepSeek-V3, deepseek-ai/DeepSeek-R1, meta-llama/Llama-3.3-70B-Instruct-Turbo, and Qwen/QwQ-32B-Preview. Note that Together AI has since updated DeepSeek-V3 to V3.1 and retired QwQ-32B-Preview.

\paragraph{Reasoning methods}
Additionally, we evaluate 9 distinct agentic reasoning approaches spanning three categories. The first category includes baseline prompting methods such as Chain-of-Thought and Self-Consistency. The second encompasses advanced prompting techniques: Multi-Persona, Self-Refine, and MedPrompt. The third category covers agent-based frameworks, including medical-specific collaboration frameworks (MedAgents and MDAgents, which we reimplemented to ensure accurate parsing of multiple-choice responses) and search-based agent methods (SPO and AFlow) with search strategies consistent with the original setting. Because agentic pipelines require extensive multi-step reasoning, evaluating them on all existing medical benchmarks would be computationally prohibitive. Therefore, Table~\ref{tab:medbench_methods} reports agentic results solely on \Ours{}, while Table~\ref{tab:medbench_dataset} provides cross-benchmark comparisons for single-call model evaluations. Our implementations of these baseline methods and evaluation framework are publicly available at \url{https://github.com/gersteinlab/MedicalAgentsBench/}.

Our data contamination analysis (Figure \ref{fig:data_contamination}) revealed that OpenAI models demonstrate consistently lower memorization metrics across all datasets compared to other model families. This finding guided our decision to use GPT-4o, GPT-4o-mini, and o3-mini as primary models for method comparison to minimize performance advantages stemming from potential training data contamination.

\paragraph{Standardized evaluation} All experiments utilize identical prompt templates and evaluation protocols to ensure a fair comparison. We implement a standardized two-round inference protocol per query for agent-based methods requiring multiple inference rounds (e.g., MedAgents). Multi-agent approaches (e.g., MDAgents) consistently employ three distinct agent roles for all evaluations.
This standardization mitigates confounding factors arising from implementation variations, allowing us to more accurately attribute performance differences to the methods themselves rather than to differences in their implementations.

\subsection*{Cost analysis}

To analyze cost-performance trade-offs, we followed a standardized evaluation protocol.
Understanding these trade-offs is particularly important given the significant computational resources required by advanced reasoning models, which often come with substantially higher inference costs and longer processing times than traditional models. Similarly, multi-agent frameworks require multiple rounds of API calls for agent interactions, further increasing computational costs and inference time.
In this analysis, we focus exclusively on inference-time cost and latency, as training costs are highly model-dependent, often proprietary, and difficult to attribute fairly on a per-task or per-sample basis.

For API-based commercial models (OpenAI and Claude), we calculated costs using their published pricing rates based on total token usage (input + output). For open-source models (such as DeepSeek, Llama, and QwQ), we estimated costs based on Together AI's hosting rates (\url{https://www.together.ai/}), a platform that provides API access to open-source models. This approach enables fair cost comparisons between proprietary and open-source alternatives across different deployment scenarios.
The total cost of experimentation was \$226.17. We measured inference time as wall-clock time per sample, including prompt construction and model inference, across agent-based methods that included their complete interaction cycles.

\section*{RESOURCE AVAILABILITY}

\subsection*{Lead contact}
Further information and requests for resources should be directed to and will be fulfilled by the lead contact, Mark Gerstein (pi@gersteinlab.org).

\subsection*{Materials availability}
This study did not generate new physical materials.

\subsection*{Data and code availability}
The benchmark dataset and all original code reported in this study are publicly available. The dataset is deposited at Zenodo \cite{shao2026medicalagentsbench_data} (\url{https://doi.org/10.5281/zenodo.18944609}) and Hugging Face (\url{https://huggingface.co/datasets/super-dainiu/MedicalAgentsBench}). The code is deposited at Zenodo \cite{shao2026medicalagentsbench_code} (\url{https://doi.org/10.5281/zenodo.18944554}) and GitHub (\url{https://github.com/gersteinlab/MedicalAgentsBench}). Any additional information required to reanalyze the data reported in this paper is available from the lead contact upon request.

\section*{ACKNOWLEDGMENTS}
We thank the anonymous reviewers for their constructive feedback. This work was supported by the ALW Professorship Funds and the National Institutes of Health Grant R01DA063148.

\section*{AUTHOR CONTRIBUTIONS}
Y.S., X.T., and J.S. conceptualized the study and methodology. Y.S. collected and curated data, developed software, conducted experiments, and created figures. J.S. curated data. J.C. and Y.L. created figures. J.Z. and J.X. conducted baseline experiments. Y.S., X.T., J.C., and J.S. wrote the manuscript. F.W., Y.Z., C.W., W.S., and A.C. reviewed and edited the manuscript. M.G. and X.T. supervised the project.

\section*{DECLARATION OF INTERESTS}
The authors declare no competing interests.

\section*{DECLARATION OF GENERATIVE AI AND AI-ASSISTED TECHNOLOGIES}

During the preparation of this work, the author(s) used Claude (Anthropic) in order to assist with manuscript formatting, editing, and bibliography management. After using this tool/service, the author(s) reviewed and edited the content as needed and take(s) full responsibility for the content of the published article.

\bibliography{custom}

\clearpage
\section*{FIGURES}

\clearpage
\section*{TABLES}

\begin{table*}[t!]
\centering
\caption{\textbf{Methods Overview.} The reasoning approaches span four categories: traditional prompting, advanced prompting, discussion-based agent frameworks, and agentic workflow optimization methods.}
\setlength{\tabcolsep}{2.8mm}{
\resizebox{\linewidth}{!}{
\begin{tabular}{lr}
\toprule
\textbf{Method} & \textbf{Description} \\
\midrule

\multicolumn{2}{l}{\textit{Traditional prompting}} \\
Chain-of-Thought \cite{wei2022chain}
& Prompts explicit intermediate reasoning steps before generating a final answer. \\
Self-Consistency \cite{wang2022self}
& Aggregates multiple independently sampled reasoning chains via majority voting. \\

\midrule
\multicolumn{2}{l}{\textit{Advanced prompting}} \\
Multi-Persona \cite{wang2023multipersona}
& Encourages diverse reasoning by assigning distinct personas within a single model. \\
Self-Refine \cite{madaan2024selfrefine}
& Iteratively refines outputs using model-generated self-feedback. \\

\midrule
\multicolumn{2}{l}{\textit{Discussion-based agent frameworks}} \\
MedPrompt \cite{chen2024medprompt}
& Uses structured multi-round prompting with ensemble-based answer aggregation. \\
MedAgents \cite{tang2023medagents}
& Coordinates multiple specialized agents for collaborative medical reasoning. \\
MDAgents \cite{kim2024mdagents}
& Dynamically adapts multi-agent interactions based on task complexity. \\

\midrule
\multicolumn{2}{l}{\textit{Agentic workflow optimization methods}} \\
AFlow \cite{zhang2024aflow}
& Searches over agentic workflows to optimize reasoning strategies. \\
SPO \cite{xiang2025spo}
& Performs self-supervised prompt optimization using task-level feedback. \\

\bottomrule
\end{tabular}
}}
\label{tab:medbench_methods}
\vspace{-.2cm}
\end{table*}

\begin{table*}[t!]
\small
\centering
\caption{\textbf{Medical Question-Answering Datasets.} Benchmarks cover professional medical examinations, biomedical literature and education, and general-domain reasoning tasks with medical content.}
\label{tab:medbench_dataset}
\setlength{\tabcolsep}{2.2mm}{
\scalebox{0.88}{
\begin{tabular}{lrrrr}
\toprule
\textbf{Benchmark} & \textbf{Size} & \textbf{Avg Len} & \textbf{Options} & \textbf{Description} \\
\midrule

\multicolumn{5}{l}{\textit{Professional medical exam benchmarks}} \\
MedQA \cite{jin2021medqa}
& 1273 & 167.1 & 4
& Medical licensing exam multiple-choice questions. \\
MedMCQA \cite{pal2022medmcqa}
& 2816 & 18.7 & 4
& AIIMS and NEET PG medical entrance exam questions. \\

\midrule
\multicolumn{5}{l}{\textit{Biomedical literature and educational QA benchmarks}} \\
PubMedQA \cite{jin2019pubmedqa}
& 500 & 316.1 & 3
& Biomedical research questions based on PubMed abstracts. \\
MedBullets \cite{chen2024medbullet}
& 308 & 213.1 & 5
& Educational medical questions from the MedBullets platform. \\
MedExQA \cite{kim2024medexqa}
& 935 & 19.1 & 4
& Multi-specialty medical questions with expert explanations. \\
MedXpertQA \cite{zuo2025medxpertqa}
& 2450 & 257.4 & 10
& Expert-level medical questions for understanding and reasoning. \\

\midrule
\multicolumn{5}{l}{\textit{General-domain benchmarks with medical coverage}} \\
MMLU \cite{hendrycks2020mmlu}
& 1089 & 55.9 & 4
& Multitask academic benchmark including medical subjects. \\
MMLU-Pro \cite{wang2024mmlu}
& 818 & 57.4 & 3--10
& Harder MMLU variant emphasizing deeper reasoning. \\

\midrule
\Ours{}
& 862 & 147.4 & 3--10
& High-difficulty medical reasoning subset across benchmarks. \\

\bottomrule
\end{tabular}
}}
\end{table*}

\begin{table*}[t]
  \centering
  \caption{\textbf{Performance by model and task.} Accuracy values are in percentages with standard errors, reported as mean $\pm$ std over 3 runs. The best values are highlighted in \textbf{bold}, and the second-best values are highlighted in \underline{underlined} format.}
  \resizebox{\linewidth}{!}{
  \begin{tabular}{l|c|c|c|c|c|c|c|c|c|}
    \toprule
    \textbf{Model} & \textbf{MedQA} & \textbf{PubMedQA} & \textbf{MedMCQA} & \textbf{MedBullets} & \textbf{MMLU} & \textbf{MMLU-Pro} & \textbf{MedExQA} & \textbf{MedXpert-R} & \textbf{MedXpert-U} \\
    \midrule
    \multicolumn{10}{c}{\textbf{Full Test Set}} \\
    \midrule
    GPT-4o-mini & 73.2 $\pm$ 1.2 & 76.6 $\pm$ 2.0 & 65.7 $\pm$ 0.8 & 53.9 $\pm$ 0.6 & 84.3 $\pm$ 1.7 & 57.1 $\pm$ 0.5 & 78.8 $\pm$ 2.2 & 13.7 $\pm$ 0.9 & 16.9 $\pm$ 1.1 \\
    \midrule
    GPT-4o & 88.2 $\pm$ 1.4 & \textbf{79.3 $\pm$ 1.7} & 76.3 $\pm$ 1.1 & 70.6 $\pm$ 1.4 & 90.9 $\pm$ 0.9 & 68.6 $\pm$ 1.7 & 84.8 $\pm$ 1.7 & 22.6 $\pm$ 0.6 & 23.5 $\pm$ 2.4 \\
    \midrule
    DeepSeek-V3 & 79.1 $\pm$ 1.1 & 73.7 $\pm$ 1.9 & 74.4 $\pm$ 0.7 & 60.6 $\pm$ 0.6 & 89.5 $\pm$ 1.0 & 65.1 $\pm$ 1.1 & 83.8 $\pm$ 1.6 & 19.0 $\pm$ 2.4 & 23.3 $\pm$ 2.4 \\
    \midrule
    o1-mini & 89.6 $\pm$ 1.7 & 77.2 $\pm$ 0.7 & 73.3 $\pm$ 0.6 & 73.5 $\pm$ 1.3 & 91.1 $\pm$ 2.2 & 67.4 $\pm$ 1.1 & 82.4 $\pm$ 0.8 & 29.5 $\pm$ 0.6 & 26.8 $\pm$ 2.0 \\
    \midrule
    o3-mini & \textbf{93.0 $\pm$ 0.5} & \underline{79.3 $\pm$ 1.9} & \underline{77.5 $\pm$ 2.0} & \textbf{82.2 $\pm$ 1.2} & \underline{93.5 $\pm$ 2.2} & \underline{69.7 $\pm$ 1.2} & \underline{85.6 $\pm$ 1.1} & \underline{34.0 $\pm$ 2.0} & \underline{31.8 $\pm$ 2.3} \\
    \midrule
    QwQ-32B & 78.9 $\pm$ 0.7 & 78.2 $\pm$ 2.0 & 69.8 $\pm$ 2.0 & 54.5 $\pm$ 1.5 & 87.2 $\pm$ 0.6 & 65.5 $\pm$ 0.6 & 81.4 $\pm$ 1.1 & 17.8 $\pm$ 2.3 & 17.3 $\pm$ 1.3 \\
    \midrule
    DeepSeek-R1 & \underline{91.8 $\pm$ 1.0} & 75.7 $\pm$ 1.1 & \textbf{82.2 $\pm$ 2.4} & \underline{78.8 $\pm$ 1.8} & \textbf{95.3 $\pm$ 2.1} & \textbf{79.7 $\pm$ 2.3} & \textbf{86.2 $\pm$ 2.1} & \textbf{37.4 $\pm$ 1.1} & \textbf{37.7 $\pm$ 1.0} \\
    \midrule
    Llama-3.3-70B & 76.8 $\pm$ 2.1 & 78.1 $\pm$ 0.5 & 71.3 $\pm$ 1.3 & 61.2 $\pm$ 0.7 & 85.6 $\pm$ 2.4 & 61.6 $\pm$ 1.5 & 78.9 $\pm$ 1.2 & 20.3 $\pm$ 2.4 & 22.6 $\pm$ 1.5 \\
    \midrule
    Claude-3.5-Sonnet & 78.1 $\pm$ 1.1 & 76.5 $\pm$ 1.7 & 68.7 $\pm$ 0.6 & 56.3 $\pm$ 2.3 & 86.5 $\pm$ 0.8 & 63.7 $\pm$ 2.5 & 81.3 $\pm$ 1.8 & 20.3 $\pm$ 1.0 & 24.1 $\pm$ 1.2 \\
    \midrule
    Claude-3.5-Haiku & 63.1 $\pm$ 1.8 & 73.7 $\pm$ 0.7 & 63.0 $\pm$ 1.1 & 49.2 $\pm$ 0.6 & 79.7 $\pm$ 1.9 & 57.6 $\pm$ 1.5 & 76.9 $\pm$ 1.8 & 14.1 $\pm$ 1.9 & 16.9 $\pm$ 2.4 \\
    \midrule
    \multicolumn{10}{c}{\textbf{Hard Test Set}} \\
    \midrule
    GPT-4o-mini & 21.8 $\pm$ 2.4 & 10.2 $\pm$ 1.7 & 17.1 $\pm$ 0.8 & 10.1 $\pm$ 2.2 & 12.2 $\pm$ 1.9 & 11.2 $\pm$ 2.4 & 4.1 $\pm$ 0.9 & 5.9 $\pm$ 0.9 & 5.2 $\pm$ 1.5 \\
    \midrule
    GPT-4o & 32.2 $\pm$ 1.1 & 8.8 $\pm$ 0.8 & \underline{25.2 $\pm$ 1.2} & 18.8 $\pm$ 2.1 & 24.8 $\pm$ 1.5 & 20.8 $\pm$ 0.6 & 18.1 $\pm$ 0.8 & 6.8 $\pm$ 2.4 & 5.8 $\pm$ 2.1 \\
    \midrule
    DeepSeek-V3 & 16.1 $\pm$ 0.7 & 12.2 $\pm$ 1.4 & 19.0 $\pm$ 1.5 & 13.4 $\pm$ 2.3 & 14.9 $\pm$ 1.8 & 11.9 $\pm$ 1.5 & 7.1 $\pm$ 0.9 & 6.0 $\pm$ 2.1 & 9.0 $\pm$ 2.3 \\
    \midrule
    o1-mini & \underline{49.1 $\pm$ 2.3} & 10.7 $\pm$ 0.9 & 20.8 $\pm$ 1.2 & 38.5 $\pm$ 1.0 & 31.4 $\pm$ 1.2 & 19.3 $\pm$ 1.6 & 15.3 $\pm$ 2.1 & \textbf{29.2 $\pm$ 2.5} & 13.9 $\pm$ 0.9 \\
    \midrule
    o3-mini & \textbf{52.9 $\pm$ 2.1} & \underline{16.0 $\pm$ 2.0} & 24.1 $\pm$ 0.6 & \textbf{50.3 $\pm$ 0.7} & \underline{35.9 $\pm$ 1.7} & 15.0 $\pm$ 0.6 & \underline{18.1 $\pm$ 1.2} & \underline{25.1 $\pm$ 1.8} & \underline{14.9 $\pm$ 1.4} \\
    \midrule
    QwQ-32B & 29.2 $\pm$ 1.9 & \textbf{16.2 $\pm$ 1.6} & 24.0 $\pm$ 1.5 & 12.4 $\pm$ 1.4 & 19.4 $\pm$ 0.7 & \underline{27.9 $\pm$ 1.8} & 9.8 $\pm$ 1.5 & 8.7 $\pm$ 1.0 & 6.0 $\pm$ 2.0 \\
    \midrule
    DeepSeek-R1 & 47.1 $\pm$ 0.7 & 12.7 $\pm$ 0.8 & \textbf{30.9 $\pm$ 2.1} & \underline{43.8 $\pm$ 2.2} & \textbf{43.7 $\pm$ 0.9} & \textbf{36.8 $\pm$ 1.6} & \textbf{26.0 $\pm$ 2.3} & 25.1 $\pm$ 0.7 & \textbf{26.3 $\pm$ 1.4} \\
    \midrule
    Llama-3.3-70B & 13.9 $\pm$ 2.2 & 12.9 $\pm$ 1.5 & 19.7 $\pm$ 0.9 & 17.1 $\pm$ 1.2 & 12.4 $\pm$ 1.1 & 9.8 $\pm$ 1.9 & 6.9 $\pm$ 2.4 & 9.1 $\pm$ 1.0 & 8.9 $\pm$ 1.1 \\
    \midrule
    Claude-3.5-Sonnet & 18.1 $\pm$ 0.6 & 10.1 $\pm$ 1.5 & 9.8 $\pm$ 1.1 & 9.1 $\pm$ 1.0 & 16.2 $\pm$ 1.5 & 13.7 $\pm$ 1.0 & 13.1 $\pm$ 2.0 & 8.8 $\pm$ 2.0 & 11.0 $\pm$ 1.8 \\
    \midrule
    Claude-3.5-Haiku & 13.0 $\pm$ 1.6 & 12.1 $\pm$ 2.2 & 22.7 $\pm$ 0.9 & 10.2 $\pm$ 1.7 & 11.2 $\pm$ 0.5 & 11.8 $\pm$ 1.0 & 13.1 $\pm$ 0.8 & 8.1 $\pm$ 1.3 & 6.0 $\pm$ 0.8 \\
    \bottomrule
  \end{tabular}
  }
  \label{tab:model_task_heatmap}
\end{table*}

\begin{table*}[t]
  \centering
  \caption{\textbf{Performance by task and method.} All tasks are evaluated on the Hard set. Each model is evaluated separately with all applicable methods. Accuracy values are in percentages, reported as mean $\pm$ std over 3 runs. The best values are highlighted in \textbf{bold}, and the second-best values are highlighted in \underline{underlined} format.}
  \resizebox{\linewidth}{!}{
  \begin{tabular}{l|c|c|c|c|c|c|c|c|c|c}
    \toprule
    \textbf{Method} & \textbf{MedQA} & \textbf{PubMedQA} & \textbf{MedMCQA} & \textbf{MedBullets} & \textbf{MMLU} & \textbf{MMLU-Pro} & \textbf{MedExQA} & \textbf{MedXpert-R} & \textbf{MedXpert-U} & \textbf{Average} \\
    \midrule
    \multicolumn{11}{c}{GPT-4o-mini} \\
    \midrule
    CoT & 24.7 $\pm$ 3.2 & 14.3 $\pm$ 1.5 & 23.3 $\pm$ 2.5 & 17.6 $\pm$ 0.6 & 26.0 $\pm$ 2.4 & \underline{35.7 $\pm$ 2.1} & 15.7 $\pm$ 1.5 & 7.3 $\pm$ 2.3 & 12.3 $\pm$ 2.5 & 19.7 $\pm$ 0.5 \\
    \midrule
    CoT-SC & 21.7 $\pm$ 2.1 & 13.3 $\pm$ 2.1 & 22.3 $\pm$ 2.1 & 15.7 $\pm$ 1.1 & \underline{27.9 $\pm$ 0.8} & \underline{35.7 $\pm$ 1.5} & \textbf{16.3 $\pm$ 3.1} & 7.7 $\pm$ 2.1 & \textbf{14.0 $\pm$ 1.0} & 19.4 $\pm$ 0.3 \\
    \midrule
    MultiPersona & 30.3 $\pm$ 1.5 & 16.3 $\pm$ 4.2 & 21.7 $\pm$ 0.6 & 16.1 $\pm$ 1.7 & 26.9 $\pm$ 1.6 & 35.0 $\pm$ 1.0 & 15.7 $\pm$ 2.3 & 6.7 $\pm$ 1.5 & 11.0 $\pm$ 1.0 & 20.0 $\pm$ 0.6 \\
    \midrule
    Self-Refine & \underline{31.0 $\pm$ 3.6} & 13.3 $\pm$ 3.2 & \underline{24.3 $\pm$ 0.6} & 17.6 $\pm$ 1.7 & \textbf{29.7 $\pm$ 2.1} & 32.0 $\pm$ 2.6 & \underline{16.0 $\pm$ 3.0} & 9.0 $\pm$ 2.0 & \textbf{14.0 $\pm$ 1.7} & \textbf{20.8 $\pm$ 0.9} \\
    \midrule
    MedPrompt & 21.0 $\pm$ 3.5 & 13.7 $\pm$ 3.2 & 18.3 $\pm$ 1.5 & 18.4 $\pm$ 1.3 & 22.8 $\pm$ 2.1 & \textbf{36.0 $\pm$ 3.0} & 14.3 $\pm$ 1.2 & 7.3 $\pm$ 1.5 & 10.7 $\pm$ 1.5 & 18.1 $\pm$ 0.5 \\
    \midrule
    MedAgents & 25.3 $\pm$ 2.3 & 12.3 $\pm$ 1.5 & 22.0 $\pm$ 1.0 & \underline{18.7 $\pm$ 2.8} & 24.2 $\pm$ 3.4 & 31.3 $\pm$ 2.5 & 12.0 $\pm$ 0.0 & 7.7 $\pm$ 2.1 & 10.7 $\pm$ 1.5 & 18.3 $\pm$ 0.4 \\
    \midrule
    MDAgents & 24.0 $\pm$ 1.0 & 17.0 $\pm$ 5.6 & 22.0 $\pm$ 1.3 & 14.2 $\pm$ 1.7 & 20.5 $\pm$ 2.4 & 32.0 $\pm$ 2.0 & 12.3 $\pm$ 1.5 & 5.3 $\pm$ 1.2 & 7.0 $\pm$ 2.0 & 17.2 $\pm$ 1.1 \\
    \midrule
    SPO & 17.7 $\pm$ 1.5 & \textbf{25.7 $\pm$ 2.1} & 20.7 $\pm$ 0.6 & \textbf{22.5 $\pm$ 0.0} & 22.4 $\pm$ 2.9 & 32.0 $\pm$ 1.0 & 15.0 $\pm$ 1.0 & \textbf{11.7 $\pm$ 1.2} & \underline{13.3 $\pm$ 2.1} & 20.1 $\pm$ 0.7 \\
    \midrule
    AFlow & \textbf{32.3 $\pm$ 2.1} & \underline{18.7 $\pm$ 3.5} & \textbf{28.0 $\pm$ 3.0} & 16.9 $\pm$ 1.1 & 25.6 $\pm$ 1.6 & 28.7 $\pm$ 1.5 & 10.0 $\pm$ 3.0 & \underline{10.7 $\pm$ 3.5} & 10.3 $\pm$ 3.5 & \underline{20.1 $\pm$ 2.1} \\
    \midrule
    \midrule
    \multicolumn{11}{c}{GPT-4o} \\
    \midrule
    CoT & 44.0 $\pm$ 4.4 & 11.3 $\pm$ 3.2 & \underline{31.3 $\pm$ 1.5} & 30.7 $\pm$ 2.3 & 31.1 $\pm$ 4.4 & 37.7 $\pm$ 2.3 & 21.0 $\pm$ 2.6 & 14.3 $\pm$ 2.1 & 14.0 $\pm$ 1.0 & 26.2 $\pm$ 1.6 \\
    \midrule
    CoT-SC & 42.3 $\pm$ 5.0 & 7.3 $\pm$ 1.5 & \textbf{32.7 $\pm$ 2.5} & 31.1 $\pm$ 0.6 & 31.5 $\pm$ 1.4 & \textbf{39.7 $\pm$ 4.2} & 20.7 $\pm$ 2.3 & 14.0 $\pm$ 4.6 & 13.7 $\pm$ 1.5 & 25.9 $\pm$ 0.7 \\
    \midrule
    MultiPersona & 45.0 $\pm$ 1.0 & 17.3 $\pm$ 2.1 & 29.7 $\pm$ 4.2 & \underline{34.5 $\pm$ 4.5} & 34.7 $\pm$ 2.9 & \underline{39.3 $\pm$ 2.5} & 21.0 $\pm$ 2.0 & \underline{16.7 $\pm$ 5.8} & 13.0 $\pm$ 2.6 & \underline{27.9 $\pm$ 1.3} \\
    \midrule
    Self-Refine & \underline{45.3 $\pm$ 3.8} & 15.0 $\pm$ 2.0 & 29.3 $\pm$ 4.0 & 33.3 $\pm$ 5.1 & \underline{36.1 $\pm$ 2.1} & 33.3 $\pm$ 3.1 & 20.3 $\pm$ 1.5 & 16.0 $\pm$ 1.0 & 16.7 $\pm$ 2.1 & 27.3 $\pm$ 0.3 \\
    \midrule
    MedPrompt & 37.3 $\pm$ 1.5 & 11.0 $\pm$ 3.0 & 28.0 $\pm$ 1.0 & 24.7 $\pm$ 3.0 & 32.4 $\pm$ 0.8 & 36.3 $\pm$ 3.5 & 21.0 $\pm$ 1.0 & 14.3 $\pm$ 1.5 & 13.3 $\pm$ 0.6 & 24.3 $\pm$ 0.3 \\
    \midrule
    MedAgents & 43.0 $\pm$ 1.0 & 17.7 $\pm$ 2.5 & 30.3 $\pm$ 0.6 & 29.6 $\pm$ 2.3 & 29.2 $\pm$ 4.8 & 37.7 $\pm$ 2.5 & 20.3 $\pm$ 1.5 & 13.0 $\pm$ 1.7 & 12.3 $\pm$ 2.1 & 25.9 $\pm$ 1.1 \\
    \midrule
    MDAgents & 37.3 $\pm$ 1.5 & 20.7 $\pm$ 2.5 & 27.2 $\pm$ 1.1 & 19.9 $\pm$ 5.5 & 28.8 $\pm$ 3.6 & \textbf{39.7 $\pm$ 1.5} & \underline{21.3 $\pm$ 1.2} & 10.9 $\pm$ 2.5 & 11.9 $\pm$ 1.4 & 24.2 $\pm$ 0.8 \\
    \midrule
    SPO & 32.0 $\pm$ 1.7 & \textbf{31.3 $\pm$ 1.5} & 30.0 $\pm$ 2.0 & 28.8 $\pm$ 0.6 & 34.7 $\pm$ 2.1 & 36.7 $\pm$ 0.6 & 21.0 $\pm$ 2.0 & \textbf{17.0 $\pm$ 2.0} & \underline{17.3 $\pm$ 1.5} & 27.7 $\pm$ 0.8 \\
    \midrule
    AFlow & \textbf{47.0 $\pm$ 1.0} & \underline{23.3 $\pm$ 4.6} & \underline{31.3 $\pm$ 1.5} & \textbf{36.0 $\pm$ 1.9} & \textbf{37.9 $\pm$ 3.4} & 36.0 $\pm$ 3.6 & \textbf{23.0 $\pm$ 2.6} & 14.3 $\pm$ 2.3 & \textbf{19.7 $\pm$ 1.5} & \textbf{29.8 $\pm$ 1.0} \\
    \midrule
    \midrule
    \multicolumn{11}{c}{o1-mini} \\
    \midrule
    CoT & 48.0 $\pm$ 4.4 & 13.7 $\pm$ 2.1 & 24.7 $\pm$ 2.3 & \textbf{43.8 $\pm$ 0.0} & 32.4 $\pm$ 3.2 & 29.0 $\pm$ 4.0 & 15.0 $\pm$ 3.0 & \underline{29.3 $\pm$ 4.0} & \textbf{18.3 $\pm$ 2.1} & 28.2 $\pm$ 0.9 \\
    \midrule
    CoT-SC & 49.7 $\pm$ 2.3 & 15.0 $\pm$ 2.0 & 24.7 $\pm$ 2.5 & 40.4 $\pm$ 1.9 & 30.6 $\pm$ 0.8 & 17.3 $\pm$ 1.5 & 16.3 $\pm$ 1.2 & 26.3 $\pm$ 1.2 & 16.0 $\pm$ 2.6 & 26.3 $\pm$ 0.6 \\
    \midrule
    MultiPersona & \underline{50.0 $\pm$ 2.6} & \underline{19.3 $\pm$ 2.3} & 24.7 $\pm$ 2.3 & \underline{42.7 $\pm$ 2.2} & \underline{33.8 $\pm$ 0.8} & 30.0 $\pm$ 1.0 & 17.3 $\pm$ 2.3 & \textbf{31.0 $\pm$ 5.3} & 16.0 $\pm$ 4.4 & \textbf{29.4 $\pm$ 1.5} \\
    \midrule
    Self-Refine & \textbf{50.7 $\pm$ 7.4} & \textbf{20.0 $\pm$ 5.0} & \textbf{27.3 $\pm$ 3.2} & 39.7 $\pm$ 4.7 & \textbf{33.8 $\pm$ 3.4} & 30.0 $\pm$ 5.3 & 16.0 $\pm$ 1.7 & 28.7 $\pm$ 2.5 & \underline{17.0 $\pm$ 1.7} & \underline{29.2 $\pm$ 1.1} \\
    \midrule
    MedPrompt & 42.3 $\pm$ 3.1 & 15.7 $\pm$ 1.5 & 21.0 $\pm$ 3.0 & 38.6 $\pm$ 1.3 & 28.3 $\pm$ 2.9 & \underline{33.0 $\pm$ 4.0} & \underline{18.7 $\pm$ 1.5} & 26.7 $\pm$ 1.5 & 16.3 $\pm$ 2.5 & 26.7 $\pm$ 0.5 \\
    \midrule
    MedAgents & 47.0 $\pm$ 4.4 & 18.0 $\pm$ 2.0 & \underline{25.7 $\pm$ 2.9} & 37.1 $\pm$ 1.1 & 27.4 $\pm$ 6.0 & \textbf{34.7 $\pm$ 1.5} & 14.0 $\pm$ 3.0 & 24.3 $\pm$ 2.5 & 13.3 $\pm$ 2.5 & 26.8 $\pm$ 1.4 \\
    \midrule
    MDAgents & 39.4 $\pm$ 3.4 & 18.3 $\pm$ 1.0 & 24.7 $\pm$ 0.6 & 36.9 $\pm$ 2.4 & 28.0 $\pm$ 2.9 & 32.9 $\pm$ 0.5 & \textbf{18.7 $\pm$ 1.6} & 21.2 $\pm$ 2.6 & 13.6 $\pm$ 2.3 & 26.0 $\pm$ 0.7 \\
    \midrule
    \midrule
    \multicolumn{11}{c}{o3-mini} \\
    \midrule
    CoT & 56.7 $\pm$ 2.1 & 15.0 $\pm$ 1.7 & \underline{27.7 $\pm$ 3.2} & 49.1 $\pm$ 1.3 & 38.4 $\pm$ 2.7 & \underline{40.3 $\pm$ 2.1} & 19.7 $\pm$ 1.5 & 28.0 $\pm$ 3.6 & 22.7 $\pm$ 3.2 & 33.0 $\pm$ 0.9 \\
    \midrule
    CoT-SC & 55.0 $\pm$ 1.7 & 14.7 $\pm$ 1.5 & 26.0 $\pm$ 1.0 & 50.2 $\pm$ 2.3 & 37.9 $\pm$ 0.8 & 38.3 $\pm$ 0.6 & 18.7 $\pm$ 1.2 & 28.0 $\pm$ 2.0 & 22.3 $\pm$ 2.1 & 32.3 $\pm$ 0.5 \\
    \midrule
    MultiPersona & \textbf{61.0 $\pm$ 1.0} & 16.7 $\pm$ 2.9 & 25.7 $\pm$ 2.1 & \underline{54.7 $\pm$ 0.6} & \underline{39.3 $\pm$ 2.9} & \underline{40.3 $\pm$ 2.1} & \underline{20.3 $\pm$ 3.1} & \underline{29.7 $\pm$ 2.9} & 22.7 $\pm$ 1.5 & \underline{34.5 $\pm$ 0.7} \\
    \midrule
    Self-Refine & 52.7 $\pm$ 2.1 & \underline{17.3 $\pm$ 1.5} & \underline{27.7 $\pm$ 2.5} & 50.9 $\pm$ 1.7 & 36.5 $\pm$ 0.8 & 31.7 $\pm$ 1.5 & \textbf{22.0 $\pm$ 2.0} & \textbf{32.8 $\pm$ 1.4} & 22.3 $\pm$ 1.2 & 32.7 $\pm$ 0.7 \\
    \midrule
    MedPrompt & \underline{59.7 $\pm$ 1.5} & 15.3 $\pm$ 1.2 & 27.3 $\pm$ 1.5 & \textbf{57.3 $\pm$ 4.5} & 33.8 $\pm$ 2.1 & 37.3 $\pm$ 0.6 & 18.3 $\pm$ 1.5 & \underline{29.7 $\pm$ 0.6} & \underline{23.3 $\pm$ 3.1} & 33.6 $\pm$ 0.6 \\
    \midrule
    MedAgents & 58.0 $\pm$ 4.0 & 16.3 $\pm$ 0.6 & 26.3 $\pm$ 2.5 & 53.5 $\pm$ 0.7 & 34.7 $\pm$ 1.6 & 38.0 $\pm$ 4.0 & 13.0 $\pm$ 0.0 & 28.7 $\pm$ 2.5 & 23.0 $\pm$ 2.6 & 32.4 $\pm$ 1.4 \\
    \midrule
    MDAgents & 58.5 $\pm$ 2.3 & \textbf{20.4 $\pm$ 4.6} & \textbf{29.9 $\pm$ 2.2} & 54.4 $\pm$ 1.5 & \textbf{41.1 $\pm$ 6.3} & \textbf{41.3 $\pm$ 5.0} & 18.3 $\pm$ 1.5 & 29.6 $\pm$ 0.9 & \textbf{23.5 $\pm$ 2.5} & \textbf{35.2 $\pm$ 1.1} \\
    \midrule
    \bottomrule
  \end{tabular}
  }
  \label{tab:method_task_heatmap}
\end{table*}

\clearpage
\setcounter{table}{0}
\setcounter{figure}{0}
\renewcommand{\thetable}{S\arabic{table}}
\renewcommand{\thefigure}{S\arabic{figure}}
\renewcommand{\theHtable}{supp.table.\arabic{table}}
\renewcommand{\theHfigure}{supp.figure.\arabic{figure}}

\begin{center}
{\large\textbf{Supplemental Information}}\\[2pt]
\textbf{MedicalAgentsBench for Complex Medical Reasoning: Comparing Internalized Reasoning Models versus Externalized Agent-based Frameworks}
\end{center}

\section*{Supplemental Methods}
\begin{table}[hbt!]
\centering
\tiny
\caption{\textbf{Reasoning depth classification examples.} Human-annotated reasoning trajectories illustrating each depth level, with gradient-style pink background highlighting to indicate successive reasoning layers.}
\label{tab:reasoning_depth_examples}
\begin{tabular}{p{1.5cm}p{13.5cm}}
\toprule
\textbf{Depth Level} & \textbf{Example} \\
\midrule

\textbf{1 (Direct)} &
\textbf{Dataset:} MedMCQA\newline
\textbf{Annotator-suggested step counts:} [1, 1, 1, 1]\newline
\textbf{Annotated reasoning steps:} 1.00\newline
\textbf{Question:}
Thyroglossal fistula is:\newline
\textbf{Options:}
A. Lined by squamous epithelium;
B. Causes a bulge in neck;
C. Inflammatory lesion;
D. Precancerous lesion\newline
\textbf{Reasoning:}
\hlc{pink!40}{A thyroglossal fistula typically presents as a midline neck swelling—a visible bulge—so the defining clinical feature is the neck mass rather than its epithelial lining, inflammatory behavior, or precancerous potential.}\newline
\textbf{Answer:}
B (Causes a bulge in neck)
\\
\midrule

\textbf{2 (Single-Step)} &
\textbf{Dataset:} MedMCQA \newline
\textbf{Annotator-suggested step counts:} [2, 1, 2, 1]\newline
\textbf{Annotated reasoning steps:} 1.50\newline
\textbf{Question:} Minimum number of lobes require to form a tooth\newline
\textbf{Options:} A. 1; B. 2; C. 3; D. 4\newline
\textbf{Reasoning:}
\hlc{pink!20}{First, we recognize that the formation of a tooth begins with the development of the tooth bud, in which the enamel organ is the key component. The enamel organ originates from a dental lamina and initially forms as a single lobe, which provides the basic structure necessary for tooth development.}\newline
\hlc{pink!60}{Second, while more complex teeth may develop additional lobes to form detailed features (such as for multiple cusp patterns in molars), the minimal requirement to form a tooth is just the one primary lobe from which all other tooth structures can subsequently differentiate.}\newline
\textbf{Answer:} A (1)
\\
\midrule

\textbf{3 (Two-Step)} &
\textbf{Dataset:} MedQA\newline
\textbf{Annotator-suggested step counts:} [3, 3, 3, 3]\newline
\textbf{Annotated reasoning steps:} 3.00\newline
\textbf{Question:}
A 37-year-old man comes to the emergency department with the chief complaint of a high fever for several days. In addition to the fever, he has had malaise, chest pain, and a dry cough. He recently went on vacation to South America but has returned to his job delivering packages. He has several friends who recently had influenza. Hi temperature is 102.8°F (39.3 °C., blood pressure is 137/80 mmHg, pulse is 104/min, respirations are 19/min, and oxygen saturation is 98\%. Chest exam reveals a deep noise found bilaterally in the lung bases. Chest radiograph reveals a wider area of opacity near the heart and bilateral lung infiltrates. Which of the following is characteristic of the most likely organism responsible for this patient's symptoms?\newline
\textbf{Options:}
A. Cultured on charcoal and yeast;
B. D-glutamate capsule;
C. Found in desert sand;
D. Polyribosyl-ribitol-phosphate capsule\newline
\textbf{Reasoning:}
\hlc{pink!20}{The patient’s presentation (high fever, malaise, chest pain, dry cough, lung findings, and bilateral infiltrates on imaging) is most consistent with pneumonia.}\newline
\hlc{pink!50}{Among the typical organisms causing community-acquired pneumonia, Streptococcus pneumoniae is the most common, and its key virulence factor is its D-glutamate capsule.}\newline
\hlc{pink!80}{Reviewing the answer options, the characteristic described for the likely organism (S. pneumoniae) is the D-glutamate capsule.}\newline
\textbf{Answer:} B (D-glutamate capsule)
\\
\midrule

\textbf{4 (Multi-Step)} &
\textbf{Dataset:} MedQA\newline
\textbf{Annotator-suggested step counts:} [4, 4, 4, 4]\newline
\textbf{Annotated reasoning steps:} 4.00\newline
\textbf{Question:}
An elderly man presents to his physician with complaints of difficulty breathing, easy fatigability, and bilateral leg swelling which began 2 months ago. His breathlessness worsens while walking, climbing the stairs, and lying flat on his back. He also finds it difficult to sleep well at night, as he often wakes up to catch his breath. His pulse is 98/min and blood pressure is 114/90 mm Hg. On examination, he has mild respiratory distress, distended neck veins, and bilateral pitting edema is evident on the lower third of his legs. His respiratory rate is 33/min, SpO2 is 93\% in room air, and coarse crepitations are heard over the lung bases. On auscultation, the P2 component of his second heart sound is heard loudest at the second left intercostal space, and an S3 gallop rhythm is heard at the apex. Medication is prescribed for his symptoms which changes his cardiac physiology as depicted with the dashed line recorded post-medication. What is the mechanism of action of the prescribed medication?\newline
\textbf{Options:}A. Decrease in transmembrane sodium gradient;
B. Preferential dilatation of capacitance vessels;
C. Inhibition of aldosterone-mediated sodium reabsorption;
D. Reduction in myocardial contractility\newline
\textbf{Reasoning:}
\hlc{pink!10}{First, we recognize the patient’s clinical presentation—dyspnea on exertion, orthopnea, paroxysmal nocturnal dyspnea, and evidence of fluid overload (distended neck veins, pitting edema, coarse basal crepitations)—indicating congestive heart failure.}\newline
\hlc{pink!30}{The cardiovascular examination findings (loud P2 at the second left intercostal space and an S3 gallop at the apex) further support the diagnosis of heart failure, with pulmonary hypertension and increased filling pressures.}\newline
\hlc{pink!50}{Given the patient’s symptoms and the fact that the medication altered cardiac physiology as shown by a dashed line in a hemodynamic graph, we infer that the prescribed treatment aimed at reducing volume overload; loop diuretics are standard in such scenarios.}\newline
\hlc{pink!70}{Loop diuretics exert their effect by inhibiting the Na\textsuperscript{+}/K\textsuperscript{+}/2Cl\textsuperscript{-} transporter in the thick ascending limb of the loop of Henle, which decreases the transmembrane sodium gradient and promotes natriuresis and diuresis, reducing preload and relieving heart failure symptoms.}\newline
\textbf{Answer:}
A (Decrease in transmembrane sodium gradient)
\\
\midrule

\textbf{5+ (Complex)} &
\textbf{Dataset:} MedQA\newline
\textbf{Annotator-suggested step counts:} [5, 5, 5, 5]\newline
\textbf{Annotated reasoning steps:} 5.00\newline
\textbf{Question:}
A 30-year-old man presents to his primary care physician with complaints of excessive fatigue and weakness for the last several weeks. He also complains of abdominal pain and constipation for the same duration. On further questioning, he reports that he has lost 8 pounds in the last 2 weeks. Past medical history and family history are insignificant. His temperature is 37.3° C (99.2° F), respirations are 21/min, pulse is 63/min, and blood pressure is 99/70 mm Hg. On physical examination, he is a tired-appearing, thin male. He has a bronze discoloration to his skin, but he denies being outside in the sun or any history of laying in tanning beds. What is the next best step in the management of this patient?\newline
\textbf{Options:}\newline
A. Administer intravenous fludrocortisone\newline
B. Start him on androgen replacement treatment\newline
C. Start him on both hydrocortisone and fludrocortisone therapy\newline
D. Order an ACTH stimulation test\newline
\textbf{Reasoning:}
\hlc{pink!0}{The patient’s symptoms (fatigue, weakness, abdominal pain, constipation, and weight loss), along with low blood pressure, strongly suggest adrenal insufficiency.}\newline
\hlc{pink!20}{The bronze skin discoloration is a classic finding in primary adrenal insufficiency due to increased ACTH-stimulating melanocytes.}\newline
\hlc{pink!40}{The clinical picture aligns with Addison's disease, a condition in which both glucocorticoid and mineralocorticoid production are deficient.}\newline
\hlc{pink!60}{In a patient with these signs and symptoms, once primary adrenal insufficiency is strongly suspected, initiating hormone replacement therapy promptly is essential.}\newline
\hlc{pink!80}{Therefore, treatment should consist of both hydrocortisone (to replace glucocorticoids) and fludrocortisone (to replace mineralocorticoids).}\newline
\textbf{Answer:}
C (Start him on both hydrocortisone and fludrocortisone therapy)
\\
\bottomrule
\end{tabular}
\end{table}
\clearpage
\begin{table}[hbt!]
\centering
\scriptsize
\caption{\textbf{Medical question quality control.} Medical student annotators evaluated questions by identifying and removing problematic items from the benchmark. Most examples fall into three main categories below.}
\label{tab:exclusion_criteria}
\begin{tabular}{p{2cm}p{13cm}}
\toprule
\textbf{Problems} & \textbf{Example} \\
\midrule

The initial answer given in the dataset was incorrect. &
\textbf{Dataset:} MedExQA (speech\_pathologist)\newline
\textbf{Question:} ``Which cranial nerve is associated with chewing?''\newline
\textbf{Options:} A. CN V; B. CN IX; C. CN X; D. CN XI.\newline
\textbf{Answer:} \hlc{pink!60}{B. CN IX (glossopharyngeal nerve)} \newline
\textbf{Issue:} Mastication is mediated by the \emph{trigeminal nerve} (CN V), whose mandibular branch innervates the major muscles of chewing (masseter, temporalis, medial/lateral pterygoids). CN IX controls taste and swallowing, not mastication. Because the marked answer contradicts established medical knowledge, the item was excluded. \\
\midrule

The question itself includes information that is not provided or cannot be obtained. &
\textbf{Dataset:} MedQA \newline
\textbf{Question:} ``A 60-year-old man is referred to a dermatologist by his family physician for management of a rare case of dermatitis that has not responded to standard therapy with corticosteroids. The patient’s medical history is unremarkable, and he is currently working reduced hours at his job as an accountant. Physical examination reveals confluent scaly patches, plaques, and generalized erythroderma along the torso and lower extremities \hlc{pink!60}{(image shown)}. There is also a solid skin lesion with a diameter greater than 1 cm. The dermatologist suspects a malignancy and orders a biopsy. Which of the following is the most accurate description of this condition?''\newline
\textbf{Options:}\newline
A. Vascular tumor containing spindle cells\newline
B. Tumor arising from the folliculosebaceous--apocrine germ\newline
C. Tumor arising from cutaneous T cells\newline
D. Tumor arising from epidermal keratinocytes\newline
\textbf{Answer:} C. Tumor arising from cutaneous T cells\newline
\textbf{Issue:} In the original exam, the diagnosis relies critically on the dermatologic image (erythroderma and lesion morphology). When the figure is omitted in text-only form, the question loses essential diagnostic information. The question misses critical clinical information or imaging required for diagnosis; the item was excluded. \\
\midrule

The question is not relevant to the medical or clinical field. &
\textbf{Dataset:} MMLU-Pro (human\_aging, health category)\newline
\textbf{Question:} \hlc{pink!60}{``With respect to marital satisfaction, most long lasting marriages''}\newline
\textbf{Options:}\newline
A. Show lower levels during the years when children are present\newline
B. Show constant low satisfaction throughout the marriage\newline
C. Show high satisfaction through all the years of marriage\newline
D. Show a steady increase in satisfaction throughout the marriage, regardless of children's presence\newline
E. Show ups and downs throughout the marriage\newline
F. Show low satisfaction in the early years, which increases as the years pass \newline
G. Show high satisfaction only during the years when children are present\newline
H. Show high satisfaction that steadily declines as the years pass\newline
I. Show high satisfaction only after children have left the home\newline
J. Show high satisfaction in the early years, which gradually increases over time.\newline
\textbf{Answer:} A. Show lower levels during the years when children are present\newline
\textbf{Issue:} Although this item appears under a ``health / human aging'' label in MMLU-Pro, it does not involve patients, disease, diagnosis, or treatment. Instead, it probes familiarity with social and relationship psychology research on marital satisfaction trajectories. Correct answers depend on recalling patterns from the psychosocial literature rather than on applying medical or geriatric expertise. Therefore, this item was excluded. \\
\bottomrule
\end{tabular}
\end{table}

\clearpage
\begin{table*}[h]
\centering
\caption{Inter-annotator agreement by dataset using Fleiss' $\kappa$.}
\setlength{\tabcolsep}{6pt}
\begin{tabular}{lclclc}
\toprule
\textbf{Dataset} & $\boldsymbol{\kappa}$ &
\textbf{Dataset} & $\boldsymbol{\kappa}$ &
\textbf{Dataset} & $\boldsymbol{\kappa}$ \\
\midrule
MedQA        & 0.73 & MedMCQA      & 0.72 & PubMedQA     & 0.65 \\
MedBullets   & 0.11 & MMLU         & 0.64 & MMLU-Pro     & 0.55 \\
MedExQA      & 0.54 & MedXpertQA-R & 0.58 & MedXpertQA-U & 0.42 \\
\midrule
\textbf{Overall} & \multicolumn{5}{c}{\textbf{0.62}} \\
\bottomrule
\end{tabular}
\label{tab:kappa_datasets}
\end{table*}
\clearpage
\section*{Supplemental Note S1. Error Analysis}
\label{app:error_analysis}

We analyze five representative clinical cases (Tables \ref{tab:medbullets_example_2}--\ref{tab:medbullets_example_5}) to characterize failure modes and success patterns across reasoning paradigms. Each table presents side-by-side reasoning traces from base models, reasoning-optimized models, and agent-based systems. Red-highlighted segments mark critical decision pivots where hypotheses are introduced, overturned, or revised.

\paragraph{Case 1 (Table \ref{tab:medbullets_example_2}): Convergent reasoning in IRIS management.}

An HIV patient on ART develops worsening pulmonary symptoms despite rising CD4 counts, a classic presentation of immune reconstitution inflammatory syndrome (IRIS). GPT-4o-mini and GPT-4o both mismanage this case: the former favors drug sensitivity testing, while the latter considers discontinuing therapy. DeepSeek-R1, by contrast, entertains alternative diagnoses (infection relapse, drug resistance) during its extended chain-of-thought but systematically eliminates each against the clinical timeline (symptom onset 4 weeks post-ART, improving immunologic markers). Each re-evaluation step narrows the differential rather than broadening it, and the model correctly selects continuation of current therapy (A). The key observation is that deliberation length per se does not determine accuracy; what matters is whether successive reasoning steps converge toward or diverge from the strongest hypothesis.

\paragraph{Case 2 (Table \ref{tab:medbullets_example_3}): Refinement versus consensus amplification in osteomyelitis.}

A patient presents with osteomyelitis following a cat bite. The clinically relevant pathogen is \textit{Pasteurella multocida}, which requires ampicillin--sulbactam (B). The initial chain-of-thought selects vancomycin (E), overweighting generic severity cues (osteomyelitis suggesting MRSA) while neglecting the pathogen-specific exposure history (cat bite indicating \textit{Pasteurella}).

Under Self-Refine, the model re-examines pathogen likelihood across refinement rounds and shifts from broad MRSA coverage to pathogen-directed therapy, ultimately converging on the correct answer. Under MultiPersona, however, three independent agents generate similar MRSA-focused rationales. During aggregation, inter-agent agreement reinforces the shared error rather than correcting it. This distinction is clinically consequential: iterative self-critique recovers from anchoring bias by forcing re-examination of premises, whereas naive aggregation amplifies correlated mistakes when agents share the same reasoning blind spot.

\paragraph{Case 3 (Table \ref{tab:medbullets_example_1}): Hypothesis reversal in small bowel obstruction.}

A patient with prior abdominal surgery and diabetes presents with symptoms consistent with adhesive small bowel obstruction. GPT-4o and o3-mini correctly identify adhesions (A) through straightforward pattern matching (surgical history + obstructive symptoms). DeepSeek-R1 initially reaches the same conclusion but then revisits the differential: the model notes the patient's diabetes and explores autonomic neuropathy (B) as an alternative etiology for dysmotility. Without an explicit stopping criterion or confidence threshold, this secondary hypothesis displaces the original, despite weaker supporting evidence.

The failure mechanism is distinct from knowledge gaps or reasoning errors in the traditional sense. The model possesses sufficient clinical knowledge to reach the correct answer and does so early in its reasoning chain. The error arises from unconstrained hypothesis exploration: additional deliberation destabilizes an already-correct conclusion. Compared with Case 1, where re-evaluation converges, this case shows that the same extended reasoning process can diverge when intermediate steps introduce plausible but weaker alternatives without calibrating their relative likelihood.

\paragraph{Case 4 (Table \ref{tab:medbullets_example_4}): Disagreement-driven correction in septic shock.}

A patient in refractory septic shock requires vasopressor escalation. The single-model chain-of-thought selects hydrocortisone (C), reasoning from persistent hypotension to adrenal insufficiency. This represents a plausible but premature inference that bypasses the standard vasopressor escalation protocol.

Within MultiPersona, agents initially disagree: some advocate corticosteroids, others argue for epinephrine as the guideline-recommended second-line vasopressor. Critically, this disagreement triggers explicit comparison of competing clinical rationales rather than simple majority voting. The synthesized response identifies that vasopressor escalation should precede steroid therapy per Surviving Sepsis guidelines, converging on epinephrine (B). The correction depends on \emph{structured} disagreement: agents with different initial hypotheses are forced to reconcile their reasoning against shared clinical criteria, exposing the premature inferential leap in the corticosteroid rationale.

\paragraph{Case 5 (Table \ref{tab:medbullets_example_5}): Role-specialized revision in neonatal galactosemia.}

A neonate presents with findings consistent with classic galactosemia. Within MedAgents, the initial synthesized report (v0) correctly identifies the diagnosis but introduces a spurious dietary recommendation (avoiding fruit juice). This is mechanistically incoherent, since galactosemia involves defective galactose metabolism, not fructose intolerance. Domain-specialized agents (ophthalmology, hepatology, clinical nutrition) flag this inconsistency: cataracts and hepatotoxicity in galactosemia are driven by galactitol and galactose-1-phosphate accumulation, not by generic sugar exposure. The revised synthesis (v1) narrows the intervention to eliminating lactose-containing feeds and switching to soy-based formula (B), correctly targeting the underlying metabolic defect. The correction arises not from agent diversity alone, but from the combination of domain-specific knowledge with an explicit revision mechanism that forces the synthesis to address identified inconsistencies.

\paragraph{Summary.}

Three recurring failure and recovery mechanisms emerge across these cases:

\begin{enumerate}
\item \textbf{Convergence versus divergence in extended reasoning.} Longer chain-of-thought improves accuracy when successive steps narrow the differential (Case 1), but degrades accuracy when unconstrained exploration displaces well-supported hypotheses (Case 3). The outcome depends on whether intermediate re-evaluation is convergent or diffusive.
\item \textbf{Consensus amplification.} Multi-agent aggregation without structured disagreement handling reinforces shared reasoning errors (Case 2, MultiPersona). When agents share the same anchoring bias, agreement increases confidence in the wrong answer.
\item \textbf{Structured correction.} Both iterative self-critique (Case 2, Self-Refine) and role-specialized disagreement resolution (Cases 4--5) can recover from reasoning errors, but only when the correction mechanism forces re-examination of specific premises rather than generating independent rationales from the same prior assumptions.
\end{enumerate}

These patterns are consistent with our quantitative findings: stronger base reasoning reduces the marginal value of agentic scaffolding, but targeted correction mechanisms (disagreement detection, premise re-examination, domain-specialized review) remain beneficial when they address specific failure modes rather than adding undifferentiated deliberation.

\begin{table*}[hbt!]
\centering
\tiny
\caption{\textbf{Example of reasoning-driven success.}
The reasoning model, e.g., Deepseek-R1, considers worsening symptoms in context, resists premature treatment changes, and maintains the strongest evidence, avoiding overreaction and producing the correct answer.
}
\label{tab:medbullets_example_2}
\begin{tabular}{p{0.32\textwidth}p{0.32\textwidth}p{0.32\textwidth}}
\toprule
\multicolumn{3}{p{\linewidth}}{
\textbf{Dataset:} MedBullets \newline
\textbf{Question:} A 31-year-old man presents with fever, malaise, cough, and hemoptysis. He was previously diagnosed with HIV (CD4 40/mm\textsuperscript{3}, viral load 115{,}000 copies/mL) and treated for a right upper lobe infiltrate. After 4 weeks on ART, his CD4 rose to >400/mm\textsuperscript{3} and viral load became negligible. Now he presents with new left upper lobe nodules and hilar adenopathy on chest radiograph. What is the most appropriate next step in management? \newline
\textbf{Options:} A. Continue current anti-retroviral treatment, antimicrobial, and anti-tuberculoid therapies; B. Determine drug sensitivities; C. Discontinue antimicrobial treatment; D. Obtain a D-dimer; E. Start high-dose glucocorticoid treatment and discontinue antiretroviral therapy. \newline
\textbf{Correct Answer:} A (Continue current therapies). \newline
\textbf{Clinical Explanation:} This patient with recently diagnosed HIV and tuberculosis infection (upper lung lobe infiltrate) is presenting with worsening symptoms after being appropriately treated. Worsening of symptoms after initiation of these therapies is concerning for immune reconstitution inflammatory syndrome (IRIS), which is managed by continuing current combined antiretroviral medications, antibiotics for opportunistic infections and symptom management. IRIS describes a paradoxical worsening of a patient's underlying infection after initiating antiretroviral therapy for patients with HIV. IRIS can also occur in HIV-uninfected patients with tuberculosis or leprosy who are started on antimycobacterial treatment. The pathogenesis of IRIS is unclear, but in patients with HIV, antiretroviral therapy increases the amount of CD4+ T-helper cells, which can lead to an exaggerated inflammatory response (e.g., increased cytokine release) towards the infection. Thus, there is a paradoxical worsening of the infection symptoms. When patients have a paradoxical worsening of their symptoms in the setting of starting their antiretroviral therapy, the therapy should be continued with symptomatic management (acetaminophen for fever). Nelson et al review IRIS. They note it is seen in patients with HIV. They recommend antiretroviral therapy and antibiotics for bacterial infections.
} \\
\midrule
\textbf{GPT-4o-mini} & \textbf{GPT-4o} & \textbf{DeepSeek-R1} \\
\midrule

B (Incorrect) &
E (Incorrect) &
A (correct) \\
\midrule
To approach this clinical scenario, we need to analyze the patient's presentation and history step-by-step:

1. \textbf{Patient Background}: The patient is a 31-year-old man with a history of HIV, presenting with fever, malaise, cough, and hemoptysis (blood in sputum). He has a very low CD4 count (40/mm\textsuperscript{3}) and a high HIV viral load (115,000 copies/mL) at the time of his previous visit.

2. \textbf{Previous Treatment}: He was treated for a right upper lobe lung infiltrate, which suggests a possible opportunistic infection, common in patients with low CD4 counts. After treatment, his CD4 count improved significantly to over 400/mm\textsuperscript{3}, and his HIV viral load became negligible.

3. \textbf{Current Symptoms}: The patient now presents with worsening cough, fever, and new findings on chest radiograph, which shows new nodules in the left upper lobe and hilar adenopathy. This raises concern for a possible new pulmonary infection or a reactivation of a previous infection, especially given his history of low CD4 count.

4. \textbf{Differential Diagnosis}: Given the patient's immunocompromised state, the differential diagnosis could include:
   - Tuberculosis
   - Fungal infections
   - Bacterial pneumonia
   - Lymphoma or other malignancies

5. \textbf{Next Steps in Management:} 
\hlc{pink!60}{The most appropriate next step in management would be to determine the cause of the new pulmonary findings. This could involve obtaining cultures or sensitivities to identify the specific pathogen responsible for the patient's symptoms.}

\textbf{Now, let's evaluate the options:}  

\hlc{pink!60}{(A) Continuing current treatments without further investigation may not address the new findings.}

\hlc{pink!60}{(B) Determining drug sensitivities against the patient’s pulmonary infection is crucial to guide appropriate treatment.}

(C) Discontinuing antimicrobial treatment is not advisable given the new symptoms.  

(D) Obtaining a D-dimer level is not relevant in this context.  

(E) Starting high-dose glucocorticoid treatment and discontinuing antiretroviral therapy is inappropriate.

Based on this analysis, the most appropriate next step in management is:  
\textbf{B) Determine drug sensitivities against the patient’s pulmonary infection.}
&

To determine the most appropriate next step in management for this patient, let's analyze the clinical scenario step-by-step:

1. \textbf{Patient History and Symptoms}: 
- 31-year-old HIV-positive man with fever, malaise, hemoptysis  
- prior CD4 40/mm\textsuperscript{3}, high viral load  
- after treatment, CD4 >400/mm\textsuperscript{3}, viral load negligible  

2. \textbf{Current Presentation}: 
- Fever 102°F  
- Nodules in left upper lobe  
- Hilar adenopathy  

3. \textbf{Possible Diagnoses}:  
\hlc{pink!60}{Given the patient's history of HIV and recent improvement in CD4 count, the new pulmonary findings could suggest an immune reconstitution inflammatory syndrome (IRIS).}  
\hlc{pink!60}{The presence of new nodules and hilar adenopathy could indicate a reactivation or worsening of a previous infection.}

4. \textbf{Management Considerations}:  
\hlc{pink!60}{Continuing current treatments (Option A) might not address the new findings adequately.}  
\hlc{pink!60}{Determining drug sensitivities (Option B) could be useful if there is suspicion of drug-resistant infection.}  
\hlc{pink!60}{Starting high-dose glucocorticoid treatment and discontinuing antiretroviral therapy (Option E) could be considered if IRIS is strongly suspected.}

\hlc{pink!60}{Given the clinical context and the likelihood of IRIS, the most appropriate next step is to manage the inflammatory response while continuing to treat the underlying HIV infection. Therefore, the best option is:  
\textbf{(E) Start high-dose glucocorticoid treatment and discontinue antiretroviral therapy.}}

\hlc{pink!60}{This approach addresses the potential IRIS by reducing inflammation and temporarily halting the immune reconstitution that may be exacerbating the symptoms.}
&

\textbf{<think>}

Okay, let's tackle this question. So, the patient is a 31-year-old man with a history of HIV. He was treated for a lung infiltrate 5 weeks ago, and his CD4 count went up after treatment. Now he's back with worsening symptoms, new nodules on the chest X-ray, and hilar adenopathy.

First, I need to recall what happens when someone with HIV starts antiretroviral therapy (ART). When they start ART, especially if they had a very low CD4 count (like 40 here), there's a risk of immune reconstitution inflammatory syndrome (IRIS). IRIS occurs when the immune system starts recovering and mounts an exaggerated response to an existing infection, causing inflammation. This can make existing infections worse or cause new symptoms even if the infection was being treated.

His previous treatment probably included antibiotics for the lung infiltrate. Now, after starting ART, his CD4 count improved (from 40 to over 400), which is a good response. But now he's presenting with new nodules and adenopathy. The timing (4 weeks after starting ART) fits with IRIS. The new nodules could be a paradoxical reaction to the treatment of an infection, like tuberculosis or another opportunistic infection. Since he was treated before and now has new findings, maybe the initial infection is still present, and the immune system is reacting to it more now.

Looking at the options: 

A) Continue current treatments. If it's IRIS, continuing ART and antimicrobials is correct, but maybe add something else? But the options don't mention adding steroids here. Wait, option E says start glucocorticoids and stop ART. But stopping ART is not recommended.

\textbf{So the best answer is A.}

\textbf{</think>}
\\
\bottomrule
\end{tabular}
\end{table*}

\begin{table*}[hbt!]
\centering
\fontsize{5}{7}\selectfont
\caption{\textbf{Example of reasoning-driven self-correction.}
Self-Refine reliably overturns an initial incorrect hypothesis through iterative reevaluation, while MultiPersona reasoning amplifies early biases across agents and may even suppress emerging correct reasoning signals.}
\label{tab:medbullets_example_3}
\begin{tabular}{p{0.32\linewidth}p{0.32\linewidth}p{0.32\linewidth}}
\toprule
\multicolumn{3}{p{\linewidth}}{
\textbf{Dataset:} MedBullets \newline
\textbf{Question:} A 5-year-old girl presents to the emergency room due to difficulty walking. She has been complaining of pain in her right leg for the last few days. Her neighbor’s cat bit her last week and her parents attributed the pain to her healing bite. At the time of the bite, they cleaned the wound and irrigated it with sterile saline water from a first aid kit. She has no medical history and has never been hospitalized. Her temperature is 102.2°F (39°C), blood pressure is 118/78 mmHg, pulse is 90/min, respirations are 21/min, and pulse oximetry is 99\% on room air. The open wound remains present on the thigh with surrounding erythema and edema. MRI is consistent with osteomyelitis. Which of the following is the most appropriate next step in management?  \newline
\textbf{Options:} A. Amoxicillin and clavulanate; B. Ampicillin and sulbactam; C. Doxycycline; D. Fluconazole; E. Vancomycin. \newline
\textbf{Correct Answer:} B (Ampicillin and sulbactam). \newline
\textbf{Clinical Explanation:} This young girl with a recent history of a cat bite and difficulty walking most likely has likely Pasteurella multocida osteomyelitis. Empiric IV antibiotic therapy with ampicillin-sulbactam is the most appropriate initial management. Pasteurella multocida is a gram-negative rod that is associated with cat and dog bites. Infection can lead to cellulitis and osteomyelitis. Treatment of minor infection is with amoxicillin-clavulanate; however, patients suspected of having osteomyelitis should be treated empirically with IV antibiotics (e.g., ampicillin and sulbactam) and taken for biopsy/culture to identify the pathogen. Antibiotics can then be narrowed down accordingly. Patients will often require between 4-8 weeks of antibiotic treatment. MRSA coverage may initially be started and tapered later once cultures result. Mollitt reviews the evidence regarding the microbiology of various injuries. He discusses how Pasteurella multocida is a common complication of animal bites. He recommends early effective antibiotic treatment.
} \\
\midrule
\multicolumn{3}{p{\linewidth}}{
\textbf{Chain-of-Thought:} E (Incorrect)} \\
\multicolumn{3}{p{\linewidth}}{
To determine the most appropriate next step in management for this 5-year-old girl presenting with symptoms consistent with osteomyelitis, let's analyze the situation step-by-step:

1. \textbf{Patient Presentation}:  
The girl has difficulty walking and pain in her right leg, which started a few days after a cat bite. The wound is still open with surrounding erythema and edema, and she has a fever (102.2°F).

2. \textbf{Vital Signs}:  
Her vital signs show a fever, but her blood pressure, pulse, respirations, and oxygen saturation are within normal limits for her age.

3. \textbf{Diagnosis}:  
The MRI confirms osteomyelitis, which is an infection of the bone. This condition can occur following a penetrating injury, such as a cat bite, especially if the wound becomes infected.

4. \textbf{Common Pathogens}:  
Cat bites can introduce various bacteria, including \textit{Pasteurella multocida}, \textit{Staphylococcus aureus}, and \textit{Streptococcus} species. Osteomyelitis in children is often caused by \textit{Staphylococcus aureus}, including methicillin-resistant \textit{Staphylococcus aureus} (MRSA).

5. \textbf{Antibiotic Choice}:  
The treatment for osteomyelitis typically involves antibiotics that cover the most likely pathogens.  
\hlc{pink!60}{Given the possibility of MRSA and the severity of the infection, the reasoning incorrectly prioritizes MRSA coverage over Pasteurella coverage.}

6. \textbf{Options Analysis}:  
- \textbf{Amoxicillin and clavulanate (A)}: Effective against many bacteria, including \textit{Pasteurella multocida}, but not the best choice for MRSA.  
- \textbf{Ampicillin and sulbactam (B)}: Effective against many bacteria but not the best choice for MRSA.  
- \textbf{Doxycycline (C)}: Can treat MRSA, but not first-line in children with severe infections.  
- \textbf{Fluconazole (D)}: Antifungal; not appropriate.  
- \textbf{Vancomycin (E)}: Effective against MRSA.  

\hlc{pink!60}{Given the need to cover MRSA and the severity of the infection, the model incorrectly concludes that the most appropriate next step is vancomycin.}

\textbf{Final Answer:}  
\hlc{pink!60}{(E) Vancomycin}
}\\
\midrule
\textbf{Self-Refine | Round 1:} A (Incorrect) & \textbf{Round 2:} B (Correct) & \textbf{Round 3:} B (Correct) \\
To determine the most appropriate next step in management for this 5-year-old girl presenting with osteomyelitis following a cat bite, let's analyze the situation step by step:\newline
1. \textbf{Patient Presentation}: The girl has difficulty walking, pain in her right leg, and a history of a cat bite. She has a fever (102.2°F), and the wound shows signs of infection (erythema and edema).\newline
2. \textbf{Diagnosis}: MRI confirms osteomyelitis, which is an infection of the bone.\newline
3. \textbf{Likely Pathogens}: Cat bites can introduce various bacteria, including Pasteurella multocida, Staphylococcus aureus, and Streptococcus species. Pasteurella multocida is particularly common in cat bites.\newline
4. \textbf{Antibiotic Choice}:
- \textbf{Amoxicillin and clavulanate (A)}: Effective against Pasteurella multocida and other common pathogens in cat bites.
- \textbf{Ampicillin and sulbactam (B)}: Also effective against Pasteurella multocida and other common pathogens.
- \textbf{Doxycycline (C)}: Effective against Pasteurella multocida but not typically first-line in young children due to potential side effects on teeth and bones.
- \textbf{Fluconazole (D)}: An antifungal, not appropriate for bacterial osteomyelitis.
- \textbf{Vancomycin (E)}: Effective against MRSA and other Gram-positive bacteria but not the first choice for Pasteurella multocida.\newline
5. \textbf{Considerations}:
- The patient is a child, so the antibiotic should be safe for pediatric use.
- The antibiotic should cover Pasteurella multocida and other common pathogens from cat bites.
\hlc{pink!60}{Given these considerations, the most appropriate next step in management is to choose an antibiotic that is effective against the likely pathogens and safe for pediatric use.}\newline
\hlc{pink!60}{Both amoxicillin and clavulanate (A) and ampicillin and sulbactam (B) are suitable choices, but amoxicillin and clavulanate is more commonly used in outpatient settings and is a well-known first-line treatment for cat bites.}\newline
Therefore, the most appropriate next step in management is:
\textbf{(A) Amoxicillin and clavulanate}
&
\textbf{Review response}:
True. The solution provided is correct. Here is a step-by-step reasoning to confirm the correctness:\newline

... [Same as Round 1]\newline

The solution correctly identifies the appropriate antibiotic for treating osteomyelitis following a cat bite in a pediatric patient.\newline
\textbf{Revised response}:
To revise the solution for the given problem, let's ensure we follow the correct steps and address the feedback appropriately.\newline
1. \textbf{Patient Presentation}: The 5-year-old girl has difficulty walking, pain in her right leg, and a history of a cat bite. She has a fever (102.2°F), and the wound shows signs of infection (erythema and edema).\newline
2. \textbf{Diagnosis}: MRI confirms osteomyelitis, which is an infection of the bone.\newline
3. \textbf{Likely Pathogens}: Cat bites can introduce various bacteria, including Pasteurella multocida, Staphylococcus aureus, and Streptococcus species. Pasteurella multocida is particularly common in cat bites.\newline
4. \textbf{Antibiotic Choice}:
- \textbf{Amoxicillin and clavulanate (A)}: Effective against Pasteurella multocida and other common pathogens in cat bites.
- \textbf{Ampicillin and sulbactam (B)}: Also effective against Pasteurella multocida and other common pathogens.
- \textbf{Doxycycline (C)}: Effective against Pasteurella multocida but not typically first-line in young children due to potential side effects on teeth and bones.
- \textbf{Fluconazole (D)}: An antifungal, not appropriate for bacterial osteomyelitis.
- \textbf{Vancomycin (E)}: Effective against MRSA and other Gram-positive bacteria but not the first choice for Pasteurella multocida.\newline
5. \textbf{Considerations}:
- The patient is a child, so the antibiotic should be safe for pediatric use.
- The antibiotic should cover Pasteurella multocida and other common pathogens from cat bites.
\hlc{teal!20}{Given these considerations, the most appropriate next step in management is to choose an antibiotic that is effective against the likely pathogens and safe for pediatric use.}\newline
\hlc{teal!20}{Both amoxicillin and clavulanate (A) and ampicillin and sulbactam (B) are suitable choices, but ampicillin and sulbactam is more commonly used in inpatient settings and is a well-known first-line treatment for osteomyelitis.}\newline
Therefore, the most appropriate next step in management is:
\textbf{(B) Ampicillin and sulbactam}
&

\textbf{Review response}:
The solution provided for the given problem is correct. Here is the step-by-step reasoning:\newline

... [Same as Round 1]\newline

The solution is correct.\newline
\textbf{Revised response}:
To revise the solution for the given problem, let's ensure we follow the correct steps and address the feedback appropriately.\newline
1. \textbf{Patient Presentation}: The 5-year-old girl has difficulty walking, pain in her right leg, and a history of a cat bite. She has a fever (102.2°F), and the wound shows signs of infection (erythema and edema).\newline
2. \textbf{Diagnosis}: MRI confirms osteomyelitis, which is an infection of the bone.\newline
3. \textbf{Likely Pathogens}: Cat bites can introduce various bacteria, including Pasteurella multocida, Staphylococcus aureus, and Streptococcus species. Pasteurella multocida is particularly common in cat bites.\newline
4. \textbf{Antibiotic Choice}:
- \textbf{Amoxicillin and clavulanate (A)}: Effective against Pasteurella multocida and other common pathogens in cat bites.
- \textbf{Ampicillin and sulbactam (B)}: Also effective against Pasteurella multocida and other common pathogens.
- \textbf{Doxycycline (C)}: Effective against Pasteurella multocida but not typically first-line in young children due to potential side effects on teeth and bones.
- \textbf{Fluconazole (D)}: An antifungal, not appropriate for bacterial osteomyelitis.
- \textbf{Vancomycin (E)}: Effective against MRSA and other Gram-positive bacteria but not the first choice for Pasteurella multocida.\newline
5. \textbf{Considerations}:
- The patient is a child, so the antibiotic should be safe for pediatric use.
- The antibiotic should cover Pasteurella multocida and other common pathogens from cat bites.
\hlc{teal!20}{Given these considerations, the most appropriate next step in management is to choose an antibiotic that is effective against the likely pathogens and safe for pediatric use.}
\hlc{teal!20}{Both amoxicillin and clavulanate (A) and ampicillin and sulbactam (B) are suitable choices, but ampicillin and sulbactam is more commonly used in inpatient settings and is a well-known first-line treatment for osteomyelitis.}\newline
Therefore, the most appropriate next step in management is:
\textbf{(B) Ampicillin and sulbactam}
\\
\midrule
\textbf{MultiPersona | Round 1:} E, B, E & \textbf{Round 2:} E, E, E & \textbf{Final:} E (Incorrect) \\
- \textbf{Persona 1 (MD Student):} \hlc{pink!60}{Given the severity of the infection (osteomyelitis) and the need to cover both common pathogens from cat bites and potential MRSA, the most appropriate antibiotic should be broad-spectrum and effective against MRSA. Vancomycin is effective against MRSA, which is a common cause of osteomyelitis in children.}\newline
- \textbf{Persona 2 (Medical Professor):} \hlc{teal!20}{The text outlines the patient's symptoms and the clinical diagnosis of osteomyelitis resulting from a cat bite. It discusses various treatment options, highlighting the necessity for a broad-spectrum antibiotic that can address the common pathogens associated with cat bites, particularly in severe infections. Among the treatment options listed, Ampicillin and sulbactam is noted as effective against many relevant bacteria, making it the most suitable choice for this patient's condition.}\newline
- \textbf{Persona 3 (Medical Researcher):} \hlc{pink!60}{Given the patient's symptoms of osteomyelitis likely caused by a cat bite and the need to cover MRSA, the best antibiotic choice is one that effectively treats these severe infections. Vancomycin is specifically noted for its efficacy against MRSA and other relevant bacteria, making it the most suitable option despite the other available choices not sufficiently covering MRSA or not being appropriate for the situation.}
&
- \textbf{Persona 1 (MD Student):}  \hlc{pink!60}{The patient presents with symptoms consistent with a severe infection, specifically osteomyelitis after a cat bite, which raises the concern for MRSA. Given that vancomycin effectively covers MRSA—a crucial factor due to the severity of the infection—this makes it the most appropriate choice for management. Other antibiotics discussed do not provide coverage for MRSA, making them less suitable in this scenario.}\newline
- \textbf{Persona 2 (Medical Professor):} \hlc{pink!60}{The patient with signs of osteomyelitis and a recent cat bite requires antibiotic treatment. The analysis shows that the infection could be due to MRSA, which is common in such cases. Reviewing the treatment options, vancomycin is identified as the only antibiotic effective against MRSA and suitable for severe infections like osteomyelitis. Therefore, vancomycin is determined to be the most appropriate choice.}\newline
- \textbf{Persona 3 (Medical Researcher):} \hlc{pink!60}{The patient's condition is severe due to osteomyelitis likely caused by a cat bite, thus requiring an antibiotic that covers MRSA and potential polymicrobial infections. Vancomycin is specifically noted for its effectiveness against MRSA and its common use in severe infections like osteomyelitis. Other options do not provide adequate coverage compared to Vancomycin.}
&
\textbf{Final Thinking}\newline
\hlc{pink!60}{The case involves a 5-year-old girl displaying severe infection symptoms following a cat bite, indicating a strong suspicion for MRSA involvement. The evaluation of antibiotic options reveals that only vancomycin effectively targets MRSA, which is critical given the high risk of severe bacterial infection indicated by her clinical presentation and MRI findings. Thus, vancomycin is determined to be the most appropriate choice for treatment.}
\\
\bottomrule
\end{tabular}
\end{table*}

\begin{table*}[hbt!]
\centering
\tiny
\caption{\textbf{Example of reasoning-driven failure.}
The reasoning model, e.g., Deepseek-R1, discovers the correct diagnosis early but, due to overthinking and repeated reconsideration of weaker alternatives, ultimately rejects its strongest evidence and outputs an incorrect answer.
}
\label{tab:medbullets_example_1}
\begin{tabular}{p{0.32\textwidth}p{0.32\textwidth}p{0.32\textwidth}}
\toprule
\multicolumn{3}{p{\linewidth}}{
\textbf{Dataset:} MedBullets \newline
\textbf{Question:} A 55-year-old man presents to the emergency department with nausea and vomiting for the past week. He decided to come in when his symptoms worsened. He feels that his symptoms are exacerbated with large fatty meals and when he drinks alcohol. His wife recently returned from a cruise with symptoms of vomiting and diarrhea. The patient has a medical history of poorly managed diabetes, constipation, anxiety, dyslipidemia, and hypertension. His temperature is 99.5°F (37.5°C), blood pressure is 197/128 mmHg, pulse is 100/min, respirations are 17/min, and oxygen saturation is 95\% on room air. Physical exam reveals a systolic murmur heard loudest along the left upper sternal border. Abdominal exam reveals an obese, tympanitic and distended abdomen with a 3 cm scar in the right lower quadrant. Vascular exam reveals weak pulses in the lower extremities. Which of the following is the most likely underlying etiology of this patient's symptoms?
\newline
\textbf{Options:} A. Adhesions; B. Enteric nervous system damage; C. Impacted stool; D. Norovirus; E. Twisting of the bowel.\newline
\textbf{Correct Answer:} A (Adhesions).\newline
\textbf{Clinical Explanation:} This patient is presenting with abdominal pain and a tympanitic abdomen, which are symptoms suggestive of a small bowel obstruction. In developed countries, these typically occur secondary to adhesions. Small bowel obstructions (SBO) typically present with an absence of bowel sounds, bowel movements, and flatus, as well as abdominal distension, nausea, and vomiting. They typically occur secondary to adhesions which entrap the bowel and disrupt its normal peristalsis. Adhesions can occur after abdominal surgery or trauma as part of the normal healing process. The diagnosis of SBO can be confirmed using abdominal imaging such as a CT scan. Treatment often involves observation; however, surgical decompression of the SBO may be needed if there is a closed-loop obstruction or an obstruction that does not resolve on its own.
} \\
\midrule
\textbf{GPT-4o} & \textbf{o3-mini} & \textbf{DeepSeek-R1} \\
\midrule

A (Correct) &
A (Correct) &
B (Incorrect) \\
\midrule
\textbf{To determine the most likely underlying etiology of this patient's symptoms, let's analyze the information provided step-by-step:}

\textbf{1. Symptoms and History:}
The patient has nausea and vomiting for the past week, worsened by large fatty meals and alcohol. His wife recently had symptoms of vomiting and diarrhea after returning from a cruise, suggesting a possible infectious etiology. The patient has a history of poorly managed diabetes, constipation, anxiety, dyslipidemia, and hypertension.

\textbf{2. Vital Signs:}
Temperature: 99.5°F (37.5°C) - slightly elevated, but not indicative of a high fever. Blood pressure: 197/128 mmHg - significantly elevated, indicating hypertensive urgency. Pulse: 100/min - tachycardia. Respirations: 17/min - normal. Oxygen saturation: 95\% on room air - normal.

\textbf{3. Physical Exam:}
Systolic murmur heard loudest along the left upper sternal border - could indicate aortic stenosis or another cardiac condition. Abdominal exam: obese, tympanitic, and distended abdomen with a 3 cm scar in the right lower quadrant - suggests previous surgery, possibly appendectomy. Vascular exam: weak pulses in the lower extremities - could indicate peripheral artery disease.

\textbf{4. Possible Etiologies:}
\textit{Adhesions (A):} Given the history of abdominal surgery (scar in the right lower quadrant), adhesions could cause bowel obstruction, leading to nausea and vomiting.
\textit{Enteric nervous system damage (B):} This could be related to his poorly managed diabetes, but it is less likely to cause acute symptoms exacerbated by fatty meals and alcohol.
\textit{Impacted stool (C):} Constipation is part of his medical history, but it is less likely to cause acute exacerbation with fatty meals and alcohol.
\textit{Norovirus (D):} The patient's wife had symptoms of vomiting and diarrhea after a cruise, suggesting a possible infectious cause. However, the patient's symptoms are exacerbated by fatty meals and alcohol, which is less typical for norovirus.
\textit{Twisting of the bowel (E):} This could cause acute symptoms of bowel obstruction, but there is no specific mention of acute, severe abdominal pain or signs of peritonitis.

Considering the patient's history of abdominal surgery, the nature of his symptoms, and the physical exam findings, the most likely underlying etiology is \textbf{adhesions} causing bowel obstruction.

Therefore, the correct answer is: \textbf{(A) Adhesions}.
&
\textbf{Step 1: Analyze the patient’s history and examination findings.}

The patient is a 55-year-old man with nausea and vomiting for a week. His symptoms worsen with fatty meals and alcohol. There is a notable 3 cm scar in the right lower quadrant, suggesting prior abdominal surgery. His exam reveals an obese, tympanitic, and distended abdomen.

\textbf{Step 2: Consider possible causes based on the provided options.}

\textit{(A) Adhesions:} After any abdominal surgery, adhesions can form. These are the most common cause of small bowel obstruction, which can present with nausea, vomiting, and abdominal distension.

\textit{(B) Enteric nervous system damage:} This could be related to diabetic gastroparesis. However, gastroparesis typically does not cause a tympanitic, distended abdomen, and a surgical scar points away from this diagnosis.

\textit{(C) Impacted stool:} While the patient has a history of constipation, impacted stool usually does not lead to prolonged vomiting over a week and a distended, tympanitic abdomen.

\textit{(D) Norovirus:} Although the wife had gastrointestinal symptoms from a recent cruise, a norovirus infection generally causes acute diarrhea and vomiting rather than a week-long picture with abdominal distension.

\textit{(E) Twisting of the bowel (volvulus):} This can cause obstruction, but it is more often seen in the context of malrotation or specific colonic volvulus and less likely with a history of prior abdominal surgery evident from the scar.

\textbf{Step 3: Correlate the findings.}

The presence of a previous surgical scar in the right lower quadrant (likely from an appendectomy) combined with the abdominal distension and vomiting strongly supports the diagnosis of a small bowel obstruction due to postoperative adhesions.

\textbf{Final Answer:} A.
&
\textbf{<think>}

Okay, let's try to work through this question step by step. The patient is a 55-year-old man with a week of nausea and vomiting that's worse after fatty meals and alcohol. His wife had vomiting and diarrhea from a cruise, which makes me think of possible infections like norovirus. But his other symptoms might point elsewhere.

His medical history includes poorly managed diabetes, which can lead to complications like gastroparesis due to autonomic neuropathy (enteric nervous system damage). ...

The systolic murmur along the left upper sternal border—maybe aortic stenosis? Or another heart issue. But the main issue here is the GI symptoms. Let's think about the possible causes of his nausea and vomiting exacerbated by fatty meals. That's classic for biliary issues like cholecystitis, but the question is about the underlying etiology. However, the options given don't include that. The options are adhesions, enteric nervous system damage (like gastroparesis), impacted stool, norovirus, or twisting (volvulus).

His wife's norovirus could be a red herring. He has risk factors for gastroparesis (diabetes), which causes delayed gastric emptying. Symptoms of gastroparesis include nausea, vomiting, especially after meals high in fat or fiber. Alcohol can also exacerbate it. The distended abdomen might be from delayed emptying. But then there's the tympanitic abdomen, which could be from gas due to obstruction or ileus.

He has a history of constipation. Impacted stool (option C) could lead to obstruction, causing nausea and vomiting, but the exacerbation with fatty meals and alcohol seems more like a motility issue. Adhesions (A) from prior surgery (the scar) could cause mechanical obstruction. But the presentation here isn't acute; it's been a week. Volvulus (E) would present more acutely with severe pain.

Diabetic gastroparesis (B) fits with the diabetes history and the triggers. The autonomic neuropathy damages the enteric nerves, leading to poor motility. The distended abdomen could be from gastric distension. The tympany might be from gas in the stomach.

\hlc{pink!60}{But wait, the abdominal exam says "distended and tympanitic," which is more suggestive of an ileus or obstruction. If it's an ileus, that's a motility problem.} Diabetic autonomic neuropathy can cause ileus. Impacted stool (C) would present with constipation and possible obstruction. But he's had constipation as part of his history. However, the exacerbation with fatty meals and alcohol points more towards gastroparesis.

\hlc{pink!60}{Alternatively, the high blood pressure and weak pulses in the legs could indicate atherosclerosis, maybe mesenteric ischemia, but that's not one of the options. The options given are A to E.}

So, the options again: A) Adhesions, B) Enteric nervous system damage, C) Impacted stool, D) Norovirus, E) Twisting.

The wife's norovirus (D) could be a coincidental exposure, but his symptoms are more chronic (a week) and not primarily diarrhea. His symptoms are more upper GI (nausea, vomiting) and related to meals.

\hlc{teal!20}{Adhesions (A) from prior surgery (scar in RLQ) could cause a small bowel obstruction.} Symptoms of obstruction include nausea, vomiting, distention, and tympanitic abdomen. \hlc{pink!60}{But the exacerbation with fatty meals and alcohol isn't typical for obstruction. However, if the obstruction is partial, maybe certain foods make it worse.} ...

\hlc{pink!60}{But the patient's history of diabetes and the triggers (fatty meals, alcohol) might still point more towards gastroparesis (B).} ...

Impacted stool (C) would present with constipation and possible obstipation. If he's been constipated and has a history of it, maybe fecal impaction leading to obstruction. But the exacerbation with meals might not directly relate.

So, the key factors here are the triggers (fatty meals, alcohol) and the diabetes. \hlc{pink!60}{Diabetic gastroparesis is a likely candidate.} So the underlying etiology would be enteric nervous system damage (B).

\hlc{pink!60}{But wait, the physical exam shows a distended and tympanitic abdomen. In gastroparesis, the stomach may be distended, but the tympany could be from that. However, if there's an ileus, that's also a motility issue. But the question is about the underlying cause. The diabetes leading to autonomic neuropathy (enteric nervous system damage) would be the etiology.}

\textbf{\hlc{pink!60}{So the answer is B.}}

\textbf{</think>}
\\
\bottomrule
\end{tabular}
\end{table*}

\begin{table*}[hbt!]
\centering
\tiny
\caption{\textbf{Example of reasoning-driven multi-agent correction.}
Here, multiple personas make early mistakes, yet the multi-persona framework reveals distinct correct rationales and resolves conflicts through consensus. This shows how structured aggregation reduces individual reasoning errors.
}
\label{tab:medbullets_example_4}
\begin{tabular}{p{0.32\linewidth}p{0.32\linewidth}p{0.32\linewidth}}
\toprule
\multicolumn{3}{p{\linewidth}}{
\textbf{Dataset:} MedBullets \newline
\textbf{Question:}
A 43-year-old man presents to the emergency department acutely with altered mental status. He uses intravenous drugs and is unable to offer a history. He was found unconscious in a park. His temperature is 99.0°F (37.2°C), blood pressure is 80/40 mmHg, pulse is 156/min, and respirations are 26/min. Physical exam reveals an obtunded man covered in vomit, with track marks on his arm and several skin abscesses. The patient is given 4 liters of Ringer lactate and intravenous antibiotics. As the fourth liter of fluids is finishing, the patient is noted to have jugular venous distension and an oxygen saturation of 92\% on nonrebreather. He is subsequently intubated. The patient is started on norepinephrine, and his blood pressure is subsequently found to be 85/44 mmHg. Which of the following is the most appropriate next step in management?
\newline
\textbf{Options:} A. Cardioversion; B. Epinephrine; C. Hydrocortisone; D. Normal saline; E. Phenylephrine
\newline
\textbf{Correct Answer:} B (Epinephrine)
\newline
\textbf{Clinical Explanation:} This patient who uses intravenous drugs with skin abscesses (the likely source of his infection) is hypotensive and tachycardic, with a blood pressure that is not responding to IV fluids; this is concerning for septic shock. Given his persistent hypotension, he should be started on vasopressors of which norepinephrine is first line and epinephrine is second line. Septic shock is a life-threatening, systemic bacterial infection that leads to vasodilation and endothelial dysfunction. Early management is centered on IV fluids, blood cultures, broad-spectrum antibiotics, and workup to uncover the source. If the patient's blood pressure does not respond to IV fluids (by definition, their blood pressure will not respond to fluids in septic shock), vasopressors should be started with a goal mean arterial pressure > 65 mmHg. The first line vasopressor is norepinephrine, which can be titrated up for maximal effect. If this vasopressor fails to reach blood pressure goals, then the next vasopressor is either vasopressin or epinephrine (either is acceptable). If the second line vasopressor fails to increase blood pressure, then the third line vasopressor is vasopressin or epinephrine (whichever was not used as the second line pressor). Finally, the fourth line vasopressor for septic shock is phenylephrine. Levy et al. review sepsis. They update the Surviving Sepsis campaign which gives the most up to date management guidelines of sepsis including adequate fluid resuscitation, broad-spectrum antibiotics, and vasopressors. They recommend early antibiotics and resuscitation to minimize morbidity and mortality.
} \\
\midrule
\multicolumn{3}{p{\linewidth}}{
\textbf{Chain-of-Thought:} C (Incorrect)} \\
\multicolumn{3}{p{\linewidth}}{
To determine the most appropriate next step in the management of this patient, let's analyze the situation step-by-step:\newline
1. \textbf{Patient Presentation}: The patient is a 43-year-old man with a history of intravenous drug use, presenting with altered mental status, hypotension, tachycardia, and respiratory distress. He was found unconscious and is now obtunded.\newline
2. \textbf{Vital Signs}:
- Temperature: 99.0°F (37.2°C)
- Blood Pressure: 80/40 mmHg initially, 85/44 mmHg after norepinephrine
- Pulse: 156/min
- Respirations: 26/min
- Oxygen Saturation: 92\% on nonrebreather\newline
3. \textbf{Physical Exam}: The patient has track marks and skin abscesses, indicating intravenous drug use. He is covered in vomit and is obtunded.\newline
4. \textbf{Initial Management}: The patient received 4 liters of Ringer lactate and intravenous antibiotics. Despite fluid resuscitation, he developed jugular venous distension and required intubation.\newline
5. \textbf{Current Status}: After intubation and starting norepinephrine, his blood pressure is still low (85/44 mmHg), and he has jugular venous distension, which suggests fluid overload or cardiac dysfunction.\newline
6. \textbf{ECG Findings}: The ECG (not provided here, but assuming it shows signs of tachycardia and possibly other abnormalities) needs to be considered in the context of the patient's clinical picture.\newline
7. \textbf{Differential Diagnosis}: The patient likely has septic shock given his history of intravenous drug use, skin abscesses, and response to antibiotics. The persistent hypotension despite fluids and norepinephrine suggests refractory septic shock or another underlying issue such as adrenal insufficiency.\newline
8. \textbf{Management Options}:\newline
- (A) Cardioversion: Typically used for arrhythmias like atrial fibrillation or ventricular tachycardia, not indicated here without specific arrhythmia evidence.
- (B) Epinephrine: Another vasopressor that could be considered, but the patient is already on norepinephrine.
- (C) Hydrocortisone: Used in septic shock, especially if adrenal insufficiency is suspected.
- (D) Normal saline: Additional fluids might worsen jugular venous distension and fluid overload.
- (E) Phenylephrine: Another vasopressor, but norepinephrine is already being used.\newline
\hlc{pink!60}{Given the patient's refractory hypotension despite fluids and norepinephrine, and the clinical suspicion of septic shock with possible adrenal insufficiency, the most appropriate next step is:}\newline
\hlc{pink!60}{(C) Hydrocortisone}\newline
\hlc{pink!60}{Hydrocortisone can help in cases of septic shock, particularly if there is adrenal insufficiency, which is a concern in this patient given his refractory hypotension.}\newline
Thus, the answer is (C) Hydrocortisone.
}\\
\midrule
\textbf{MultiPersona | Round 1:} C, C, B & \textbf{Round 2:} B, B, C & \textbf{Final:} B (Correct) \\
- \textbf{Persona 1 (MD Student):} \hlc{pink!60}{The patient's persistent hypotension despite norepinephrine and the presence of signs indicating fluid overload suggests that there may be adrenal insufficiency or refractory septic shock at play. Hydrocortisone is indicated in such situations as it can help address potential relative adrenal insufficiency during septic shock. Therefore, it is the most appropriate next step in management.}\newline
- \textbf{Persona 2 (Medical Professor):} 
\hlc{pink!60}{The patient is not responding adequately to fluids and norepinephrine, suggesting the possibility of adrenal insufficiency in septic shock. Hydrocortisone is appropriate for patients in septic shock who are not adequately responding to other treatments.}\newline
- \textbf{Persona 3 (Medical Researcher):} \hlc{teal!20}{Analyzing the clinical scenario, it is clear that the patient is in septic shock with persistent hypotension despite the initial treatment with fluids and norepinephrine. The options were evaluated, and while several medications were considered, epinephrine is specifically indicated for septic shock when there is inadequate response to norepinephrine. It can enhance cardiac output and systemic vascular resistance, making it the most appropriate next step in this case.}
&
- \textbf{Persona 1 (MD Student):}  
\hlc{teal!20}{The patient is experiencing refractory septic shock with persistent hypotension despite fluid resuscitation and norepinephrine. The recommendation is to use epinephrine as it can improve cardiac output and vascular resistance, which is necessary to stabilize the patient's condition.}\newline
- \textbf{Persona 2 (Medical Professor):} \hlc{teal!20}{Given the patient's persistent hypotension despite norepinephrine administration, epinephrine is identified as a reasonable next step since it enhances cardiac output and systemic vascular resistance, which is crucial in managing refractory septic shock. It is regarded as the most appropriate option for stabilization in this scenario.}\newline
- \textbf{Persona 3 (Medical Researcher):} \hlc{pink!60}{The patient presents with refractory hypotension despite fluid resuscitation and vasopressor therapy. This suggests a potential adrenal insufficiency in the context of septic shock, which requires treatment with hydrocortisone. Other options do not align with the management of the current clinical scenario.}
&
\textbf{Final Thinking}\newline
\hlc{teal!20}{The patient is in refractory septic shock with persistent hypotension despite fluid resuscitation and norepinephrine administration, indicating that initial treatment is ineffective. Given the options, epinephrine stands out as the most appropriate next step because it provides both vasopressor and inotropic support, which is crucial for stabilizing the hemodynamics in this scenario.}
\\
\bottomrule
\end{tabular}
\end{table*}

\begin{table*}[hbt!]
\centering
\tiny
\caption{\textbf{Example of MedAgents consensus-driven correction.}
Unlike naive multi-agent voting, which amplifies shared biases, MedAgents employs domain decomposition, synthesis, and cross-expert verification to identify and correct reasoning faults. Disagreements trigger revisions, enabling iterative refinement toward more reliable, clinically aligned answers.
}
\label{tab:medbullets_example_5}
\begin{tabular}{p{0.32\linewidth}p{0.32\linewidth}p{0.32\linewidth}}
\toprule
\multicolumn{3}{p{\linewidth}}{
\textbf{Dataset:} MedBullets \newline
\textbf{Question:} 
A 4-week-old boy is brought to the pediatrician by his parents for an initial evaluation. His parents are concerned that he is not feeding well and has lost weight over the last 2 weeks. He was born at home via spontaneous vaginal delivery at 37 weeks of gestation to a G1P1 mother. His mother did not receive prenatal care. His blood pressure is 70/44 mmHg, pulse is 151/min, and respirations are 41/min. His weight is 3 kg (4th percentile) compared with 3.5 kg at birth (45th percentile). On physical exam, he is jaundiced with an enlarged liver and spleen. He does not appear to respond to visual stimuli, and further examination reveals bilateral clouding of the lens. Which of the following interventions could have avoided this patient's symptoms? \newline
\textbf{Options:} A. Avoiding fruit juice and sweetened foods; B. Changing to a soy-based formula; C. Providing imiglucerase enzyme replacement; D. Removing phenylalanine from maternal diet during pregnancy; E. Vitamin B\textsubscript{6} supplementation. \newline
\textbf{Correct Answer:} B (Changing to a soy-based formula). \newline
\textbf{Clinical Explanation:} 
This patient who presents with failure to thrive, hepatosplenomegaly, and bilateral cataracts most likely has classic galactosemia. Patients with this disorder should avoid lactose-containing products by changing to a soy-based formula. Classic galactosemia is an autosomal recessive defect in galactose-1-phosphate uridyltransferase. This enzyme is involved in the conversion of galactose to glucose, and a deficiency of this enzyme results in the accumulation of galactose 1-phosphate in the liver, kidney, and brain. This metabolite acts as a phosphate sink, meaning that it traps all free phosphate in the cytosol and inhibits the formation of other phosphate-dependent metabolites such as adenosine triphosphate. This results in failure to thrive, vomiting, lethargy, and hepatosplenomegaly after ingestion of galactose-containing products such as milk. Patients will also develop bilateral cataracts if the deficit is not detected in the first few weeks after birth. Avoiding galactose by changing to a soy-based formula will avoid the complications of this disorder. Turck discusses using soy protein for infant feeding. The author notes that there are safety concerns with regards to the high phyto-estrogenic isoflavone content of these formulas. The author recommends that despite these safety concerns, these formulas should still be used in patients with classic galactosemia.
} \\
\midrule
\multicolumn{3}{p{\linewidth}}{
\textbf{MedAgents expert analyses (5 domains).} \newline
\textbf{Pediatrics Expert:} 
This 4-week-old infant shows failure to thrive (weight decline from the 45th to the 4th percentile), poor feeding, tachycardia, and hepatosplenomegaly in the setting of persistent jaundice. The bilateral lens clouding and lack of visual tracking point to early-onset cataracts rather than isolated neonatal jaundice. The subacute time course after birth and the multisystem involvement (growth, liver, and eyes) strongly suggest a congenital metabolic disorder unmasked by routine feeding rather than an isolated infectious process. The most worrisome elements are the early age, progressive weight loss, and evolving liver disease, which together indicate an ongoing toxic exposure. \newline
\textbf{Medical Genetics Expert:} 
The triad of hepatosplenomegaly, jaundice, and neonatal cataracts is highly characteristic of classic galactosemia due to GALT deficiency. Without newborn screening or prenatal care, this diagnosis may be missed until symptoms appear after exposure to lactose-containing feeds. In galactosemia, excess galactose is phosphorylated to galactose-1-phosphate and also reduced to galactitol; these metabolites disrupt hepatic function and lens transparency. The key concern is that continued galactose intake rapidly leads to liver failure, sepsis (often \textit{E.~coli}), and neurodevelopmental injury, even though these outcomes are preventable with early dietary restriction. \newline
\textbf{Neonatology Expert:} 
A term infant with poor feeding, weight loss, and hepatosplenomegaly at 4 weeks raises alarm for a serious neonatal condition. The home birth and lack of prenatal care increase the likelihood of missed newborn screening. The constellation here---failure to thrive, cholestatic-appearing jaundice, and bilateral cataracts---is most consistent with a systemic inborn error of metabolism triggered postnatally by a specific nutrient in the diet. The most alarming feature is that every feed likely delivers the toxic substrate, so the clinical trajectory can deteriorate quickly unless the offending nutrient is removed. \newline
\textbf{Ophthalmology Expert:} 
Symmetric lens clouding in a neonate is classic for metabolic cataracts, particularly those caused by galactitol accumulation in galactosemia. Aldose reductase converts excess galactose into galactitol within the lens, creating an osmotic gradient that leads to lens fiber swelling and opacification. The absence of visual tracking indicates significant functional impairment. Importantly, early recognition and removal of galactose can prevent progression and, in some cases, partially reverse cataract formation. Unchecked exposure, however, risks permanent visual deficits. \newline
\textbf{Hepatology Expert:} 
Jaundice and hepatosplenomegaly at 4 weeks reflect true hepatic dysfunction, not physiologic jaundice. In classic galactosemia, galactose-1-phosphate accumulates in hepatocytes, leading to cholestasis, synthetic dysfunction, and enlargement of the liver and spleen. If dietary galactose is not eliminated, this process can progress to acute liver failure, coagulopathy, and sepsis. The most urgent issue is that the same nutrient (lactose/galactose) driving liver injury is still present in each feed, making prompt dietary modification central to both treatment and prevention.
}\\
\midrule
\textbf{Synthesized Report (v0):} & \textbf{Vote Result (all domains):} & \textbf{Revised Synthesized Report (v1):} \\
\textbf{Key Knowledge:} Classic galactosemia due to GALT deficiency explains the combination of failure to thrive, jaundice, hepatosplenomegaly, and neonatal cataracts in a 4-week-old infant who likely missed newborn metabolic screening. Ongoing ingestion of lactose-containing feeds results in accumulation of galactose-1-phosphate in the liver and galactitol in the lens. Early dietary intervention that removes the offending sugars can prevent hepatic failure and cataract progression. \newline
\textbf{Total Analysis (v0):} 
Integrating the experts' reports, the system concludes that this infant's presentation is best accounted for by a galactose-processing defect leading to toxic accumulation of galactose metabolites. Because both galactose and other simple sugars can place metabolic stress on the immature liver, a cautious preventive strategy is to broadly limit sugar exposure. Accordingly, the initial synthesis recommends \hlc{pink!60}{avoiding fruit juice and sweetened foods to reduce additional sugar load, and switching from standard milk or formula to a lactose-free, soy-based formula to eliminate dietary galactose.} This combined approach aims to minimize exposure to both galactose and other potentially harmful simple carbohydrates, thereby preventing the development of cataracts, hepatosplenomegaly, and failure to thrive in susceptible infants. 
&
\textbf{Pediatrics Expert:} Yes. The overall diagnosis (classic galactosemia) and emphasis on removing lactose/galactose are appropriate; the mention of fruit juice has little impact at 4 weeks but is not overtly harmful. \newline
\textbf{Medical Genetics Expert:} Yes. The core mechanism and the need to eliminate galactose are correctly described, though the focus on fructose-containing juices is somewhat tangential. \newline
\textbf{Neonatology Expert:} Yes. The synthesis captures that early dietary modification is preventive, although the practical relevance of fruit-juice avoidance is limited at this age. \newline
\textbf{Ophthalmology Expert:} \hlc{pink!60}{No. The cataracts here are specifically due to galactitol accumulation from galactose, not from generic sweetened foods. The report should emphasize that prevention hinges on eliminating galactose, not on broad sugar restriction.} \newline
\textbf{Hepatology Expert:} \hlc{pink!60}{No. The liver injury is driven by galactose-1-phosphate accumulation from lactose-containing feeds. The recommendation to avoid fruit juice is clinically irrelevant for a 4-week-old and distracts from the urgent need to remove galactose.} \newline
\textbf{Clinical Nutrition Expert:} \hlc{pink!60}{No. Fruit juice and sweetened foods are not part of a typical 4-week-old diet; the preventive intervention should focus solely on changing the infant's primary milk source to a lactose-free formula.} \newline
\textbf{Outcome:} Because several specialists disagree, MedAgents triggers a revision round, asking dissenting experts to propose edits that sharpen the causal link between galactose exposure and the preventive intervention.
&
\textbf{Incorporated Revision Advice:} Ophthalmology emphasizes that cataracts arise from galactitol produced specifically from galactose, not general sugar intake. Hepatology and Clinical Nutrition stress that fruit juice is irrelevant at 4 weeks and that the key is immediate removal of galactose-containing feeds. \newline
\textbf{Total Analysis (v1):} 
After incorporating the revision advice, the updated synthesis focuses on the precise causal relationship between galactose exposure and organ damage. Classic galactosemia due to GALT deficiency leads to accumulation of galactose-1-phosphate in the liver and conversion of excess galactose to galactitol in the lens. These metabolites cause cholestatic liver disease, failure to thrive, and neonatal cataracts. Crucially, these complications are \emph{specifically triggered by continued ingestion of lactose (and thus galactose) in breast milk or standard formula}. Preventing this clinical picture does not require broad restriction of all sugars, nor is it meaningfully influenced by fruit juice in a 4-week-old infant. Instead, the central preventive measure is to \hlc{teal!20}{eliminate dietary galactose as soon as the disorder is recognized by switching the infant from breast milk or standard formula to a lactose-free, soy-based formula.} This intervention removes the toxic substrate, halts further accumulation of galactose metabolites, and can prevent or mitigate both hepatic injury and cataract formation. Alternative options such as avoiding fruit juice, using imiglucerase, restricting maternal phenylalanine, or providing vitamin B\textsubscript{6} do not address the underlying defect and would not have avoided this patient’s symptoms. \newline
\textbf{Final MedAgents Answer:} \hlc{teal!20}{B. Changing to a soy-based formula.}
\\
\bottomrule
\end{tabular}
\end{table*}

\end{document}